\documentclass[10pt,twocolumn,letterpaper]{article}

\usepackage[pagenumbers]{iccv} 

\definecolor{iccvblue}{rgb}{0.21,0.49,0.74}
\usepackage[pagebackref,breaklinks,colorlinks,allcolors=iccvblue]{hyperref}

\usepackage{multirow}
\usepackage{array}
\usepackage{amsfonts}
\usepackage{marvosym}
\usepackage{algorithm}
\usepackage{algorithmic}
\usepackage{graphicx}

\usepackage{alphalph}

\usepackage{wrapfig}

\definecolor{darkpink}{RGB}{255, 20, 147}
\usepackage{hyperref}

\def\paperID{7835} 
\def\confName{ICCV}
\def\confYear{2025}

\title{DADM: Dual Alignment of Domain and Modality for Face Anti-spoofing}

\author{
Jingyi Yang\textsuperscript{\rm 1,2}\thanks{Equal Contribution} \and
Xun Lin\textsuperscript{\rm 1 *} \and
Zitong Yu\textsuperscript{\rm 1,3}\thanks{Corresponding author} \and
Liepiao Zhang\textsuperscript{\rm 4,5} \and
Xin Liu\textsuperscript{\rm 6} \and
Hui Li\textsuperscript{\rm 2} \and
Xiaochen Yuan\textsuperscript{\rm 7} \and
Xiaochun Cao\textsuperscript{\rm 8} \and
\textsuperscript{\rm 1}Great Bay University \quad \textsuperscript{\rm 2}University of Science and Technology of China \\
\textsuperscript{\rm 3}Dongguan Key Laboratory for Intelligence and Information Technology \\ 
\textsuperscript{\rm 4}GRGBanking Equipment Co., Ltd. \quad \textsuperscript{\rm 5}South China University of Technology \\ 
\textsuperscript{\rm 6}Lappeenranta University of Technology \quad \textsuperscript{\rm 7}Macao Polytechnic University \\
\textsuperscript{\rm 8}Shenzhen Campus of Sun Yat-sen University
}

\begin{document}
\maketitle
\begin{abstract}
With the availability of diverse sensor modalities (i.e., RGB, Depth, Infrared) and the success of multi-modal learning, multi-modal face anti-spoofing (FAS) has emerged as a prominent research focus. The intuition behind it is that leveraging multiple modalities can uncover more intrinsic spoofing traces. However, this approach presents more risk of misalignment. We identify two main types of misalignment: (1) \textbf{Intra-domain modality misalignment}, where the importance of each modality varies across different attacks. For instance, certain modalities (e.g., Depth) may be non-defensive against specific attacks (e.g., 3D mask), indicating that each modality has unique strengths and weaknesses in countering particular attacks. Consequently, simple fusion strategies may fall short. (2) \textbf{Inter-domain modality misalignment}, where the introduction of additional modalities exacerbates domain shifts, potentially overshadowing the benefits of complementary fusion. To tackle (1), we propose a alignment module between modalities based on mutual information, which adaptively enhances favorable modalities while suppressing unfavorable ones. To address (2), we employ a dual alignment optimization method that aligns both sub-domain hyperplanes and modality angle margins, thereby mitigating domain gaps. Our method, dubbed \textbf{D}ual \textbf{A}lignment of \textbf{D}omain and \textbf{M}odality (DADM), achieves state-of-the-art performance in extensive experiments across four challenging protocols demonstrating its robustness in multi-modal domain generalization scenarios. Our code is available at \href{https://github.com/yjyddq/DADM}{\textcolor{darkpink}{here}}.
\end{abstract}    
\vspace{-0.7cm}
\section{Introduction}
\label{sec:introduction}

\begin{figure}[t!]
\centering
    \includegraphics[width=0.46\textwidth]{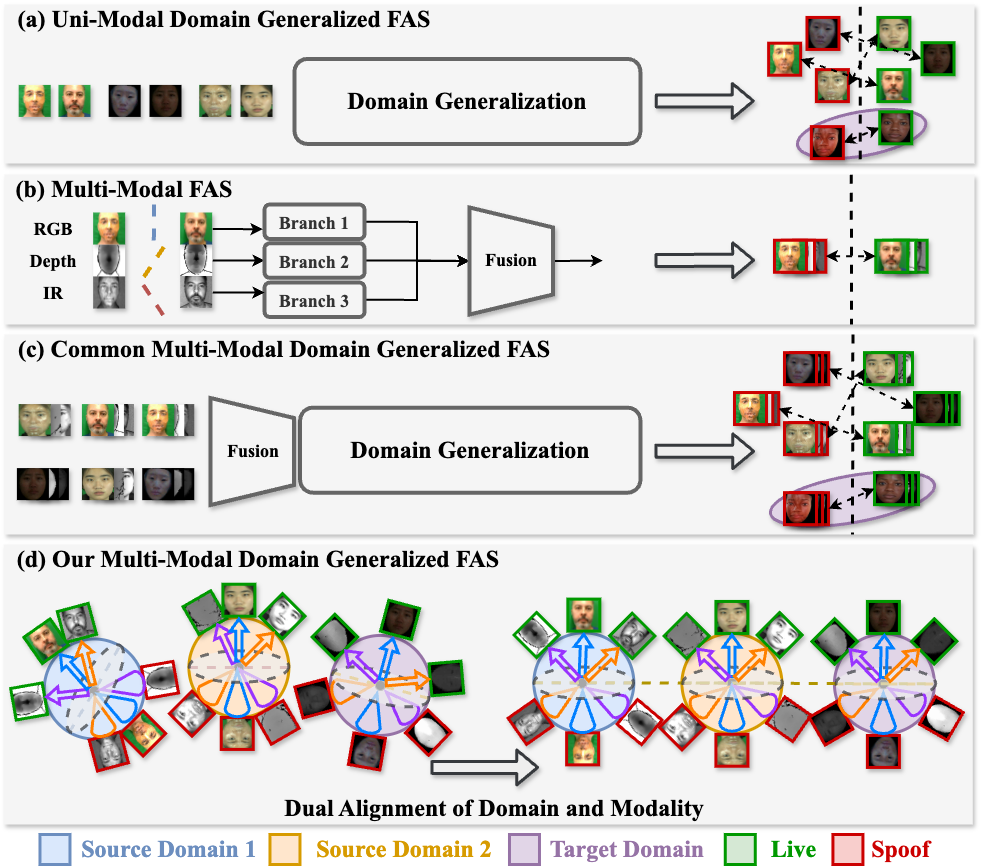}
    \vspace{-0.3cm}
    \caption{Four common scenarios of FAS. (a) Uni-modal DG-FAS aims to mitigate domain shifts. (b) Multi-modal FAS focuses on efficient fusion among modalities. (c) Common multi-modal DG-FAS, which combines DG-FAS with multi-modal fusion techniques. (d) Our proposed multi-modal DG-FAS pursues both the hyperplane alignment across each sub-domain and the angle margin consistency among modalities. For clarity, it should be noted that the different-colored spheres (domains) represent distinct faces of a sphere and do not indicate the Euclidean distance.}
    \label{fig:motivation}
    \vspace{-0.5cm}
\end{figure}

Face recognition (FR) is widely used in various applications, such as access control, phone unlocking, and mobile payments. However, FR systems are vulnerable to a wide range of presentation attacks, including but not limited to photo/paper printing, video replay, makeup, and 3D masks. To this end, face anti-spoofing (FAS) techniques have been developed to safeguard FR systems. Over the past few decades, both hand-crafted features based approaches~\cite{peixoto2011face,de2013lbp,komulainen2013context,boulkenafet2015face,boulkenafet2016face,patel2016secure} and deep learning based algorithms~\cite{yang2014learn,atoum2017face,atoum2017face,liu2018learning,yang2019face} have shown promising results in uni-modal FAS. Despite their effectiveness in intra-dataset evaluations, they generalize poorly to unseen domains (i.e., domain shifts). 

To address this issue, uni-modal domain generalization (DG) FAS methods~\cite{jia2020single,wang2022domain,li2022one,zhou2023instance,sun2023rethinking,le2024gradient,zhou2024test} have been extensively explored, as shown in Fig.~\ref{fig:motivation}~(a). Recently, with the growing challenge of attack patterns and the acquisition of more advanced sensors, FAS has expanded from uni-modal (RGB) to multi-modal (e.g., RGB, Depth, Infrared) approaches~\cite{parkin2019recognizing,george2019biometric,zhang2020casia}. These multi-modal methods aim to leverage complementary information across modalities, enabling spoof traces undetectable in one modality to be captured by others. However, existing multi-modal FAS methods often assume consistent training and testing domains, focusing primarily on modality fusion without adequately considering domain shifts, as illustrated in Fig.~\ref{fig:motivation}~(b). Furthermore, since each modality possesses varying defensive capabilities against different attacks, treating or integrating them equally may not yield optimal results. 

In multi-modal FAS, DG scenarios involve significant domain shifts~\cite{lin2024suppress} caused by advanced or unseen attacks, variations in sensor resolutions, deployment environments, and disturbance from low-quality sensors. We summarize the challenges in multi-modality domain generalization FAS into two main aspects: (1) \textbf{intra-domain modality misalignment}, where the relative importance of each modality varies for diverse attacks. Certain modalities (e.g., depth) may be vulnerable to specific attacks (e.g., 3D masks), making them unreliable. Simple fusion strategies potentially propagate negative impacts across modalities. (2) \textbf{inter-domain modality misalignment}, where incorporating additional modalities can exacerbate domain shifts. In such cases, the adverse effects of domain shifts in multi-modality may outweigh the benefits of fusion. Previous work MMDG~\cite{lin2024suppress} employs single-side prototypical loss for domain generalization, akin to conventional uni-modal DG methods that use mixed source domain, as illustrated in Fig~\ref{fig:motivation}~(c). MMDG~\cite{lin2024suppress} proposes a Monte Carlo dropout-based~\cite{carreno2023adapting,mae2021uncertainty,xixi2023uncertainty} uncertainty estimation module to recognize unreliable information. However, it has a certain degree of randomness, limiting the model’s performance and necessitating further improvements.

\begin{figure*}
\vspace{-1.8em}
  \centering
    \includegraphics[width=1.\textwidth]{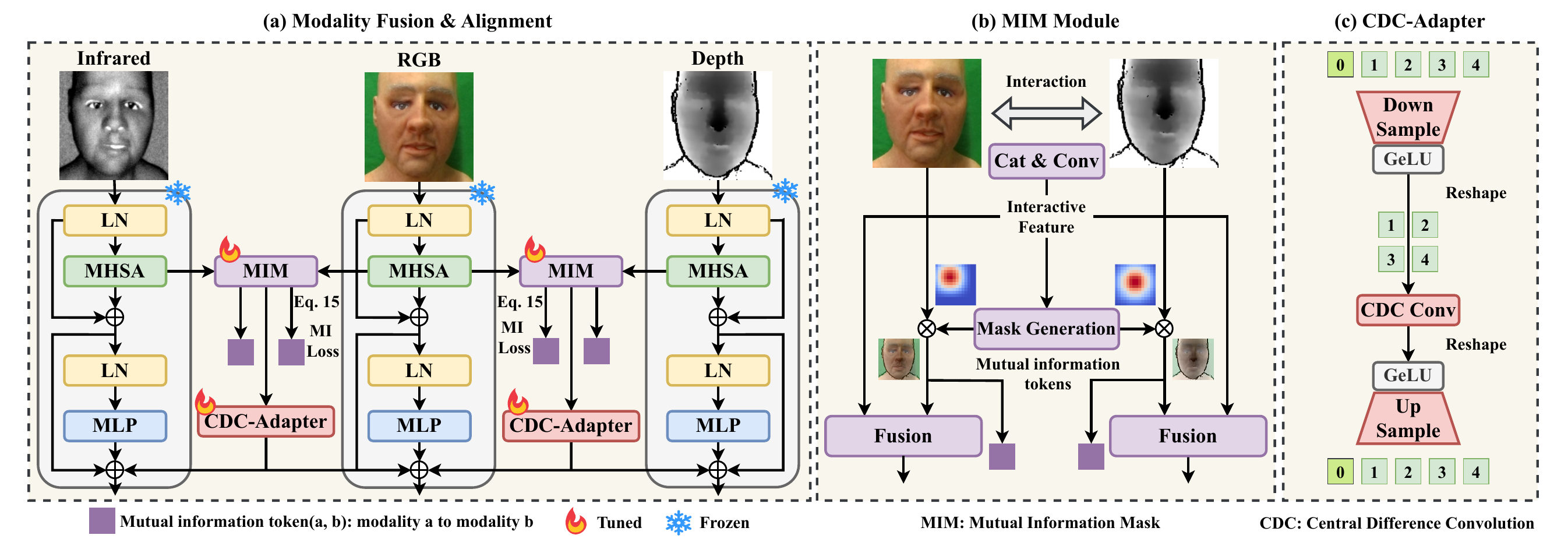}
    \vspace{-0.8cm}
    \caption{(a) IIustration of fine-tuning ViT with proposed MIM modules and CDC-Adapters, showcasing the interaction between the RGB and Depth modalities. Note that only parameters of MIM and CDC-Adapter are trainable. (b) Mutual Information Mask (MIM) module is used for suppress and enhance the informative region of features. (c) CDC-Adapter is used for integrating fine-grained local features.}
    \label{fig:architecture}
\vspace{-0.3cm}
\end{figure*}

To address these challenges, we propose the Dual Alignment of Domain and Modality (DADM) framework, as illustrated in Figs.~\ref{fig:architecture} and~\ref{fig:motivation}~(d). Specifically, we design a Mutual Information Mask (MIM) module to fine-tune ViT~\cite{dosovitskiy2020image}. The MIM module alleviates intra-domain modality misalignment through mutual information maximization. Based on the intuition that observing a certain modality can reduce the uncertainty of other modalities to some extent (i.e., we have a prior uncertainty $\mathrm{H}(X)$ for $X$ and a posterior uncertainty $\mathrm{H}(X|Y)$ given $Y$, where $\mathrm{H}(X)-\mathrm{H}(X|Y)$ is mutual information). This allows the observed modality to diminish unreliable information about other modalities and provide informative guidance. To tackle inter-domain modality misalignment, we propose a dual alignment of domain and modality optimization strategy, which aims to find a unified classification hyperplane and a unified angle margin among modalities. Our contributions are as follows:
\begin{itemize}
\item We propose the DADM framework to enhance the domain and modality generalization in multi-modal DG FAS.
\item      We introduce the Mutual Information Mask module to alleviate intra-domain modality misalignments by enhancing reliable modalities and suppressing unreliable ones.
\item  We employ a dual alignment of domain and modality optimization strategy to align sub-domain hyperplanes and modality angle margins, mitigating the inter-domain modality misalignment. 
\item We evaluate our DADM approaches under four challenging protocols. Extensive experiments demonstrate its superiority and generalization capability.
\end{itemize}

\section{Related Work}
\label{sec:related work}

\textbf{Uni\&Multi-Modal Domain Generalization in FAS.}
In recent years, uni-modal domain generalization (DG) methods for FAS have received extensively research attention~\cite{li2018domain,shao2019multi,jia2020single,shao2020regularized,chen2021generalizable,wang2022domain,zhou2023instance,sun2023rethinking,liu2023towards,yang2024generalized,le2024gradient,zhou2024test}. Common strategies include adversarial training~\cite{shao2019multi,jiang2023adversarial,liu2022spoof,yue2023cyclically}, asymmetric triplet loss~\cite{jia2020single,liao2023domain}, contrastive learning~\cite{liu2023towards}, meta-learning~\cite{shao2020regularized,cai2022learning,du2022energy,jia2021dual,qin2021meta,zhou2022adaptive}, style augmentation~\cite{wang2022domain,zhou2023instance,zhou2024test}. They share a common goal: training a model on multiple source domains, with the intention that ensure effective generalization to unseen domains. Multi-modal FAS methods have evolved in parallel. Early methods include early-fusion~\cite{george2020learning,nikisins2019domain,xie2024fusionmamba} and late-fusion~\cite{kong2022beyond,kong2024m,shen2019facebagnet,yu2020multi,george2021cross}. More recently, works that introduce attention- and ViT-based feature fusion techniques have emerged~\cite{deng2023attention,yu2023visual,yu2024rethinking,yu2023flexible,yang2024g,yang2025kronecker}. In flexible-modal FAS, cross-modality attention~\cite{liu2023ma,liu2023fm} and multi-modal adapters~\cite{yu2023visual,yu2024rethinking} enable pretrained ViTs to learn modality-agnostic features. However, despite these advancements, these methods often fall short in domain generalization scenarios due to insufficient ability to resist domain shifts. Furthermore, those DG-based approaches are proposed for uni-modal FAS and are not suitable for multi-modal ones. Additionally, a previous study~\cite{lin2024suppress} has demonstrated that the uni-modal DG-based strategies exhibit limited performance in multi-modal scenarios.
 
\noindent
\textbf{Mutual Information Neural Optimization.}
Mutual information has a wide range of applications in deep representation learning~\cite{chen2016infogan,bachman2019learning,tian2020contrastive,cheng2020club,lee2021infomax,liao2021multimodal,zhang2021rgb,tu2023implicit}. Many studies~\cite{hu2017learning,belghazi2018mutual,oord2018representation,hjelm2018learning,tschannen2019mutual} have focused on optimizing mutual information in deep learning. They often maximize the mutual information between features extracted from diverse views, modalities, or images which derived from data augmentation aims to capture high-level factors whose influence spans different perspectives - e.g., the presence of certain different perspectives of spoof trace or occurrence of certain inconsistency in data. This capability is especially valuable in multi-modal FAS, where each modality brings unique advantages or weaknesses in countering specific attack types. By maximizing mutual information between modalities, the model can adaptively emphasize task-relevant information, thereby enhancing reliable modalities while mitigating the impact of unreliable ones.

\section{Proposed Method}
\label{sec:proposed method}

Sec.~\ref{sec:preliminaries} reviews the mutual information maximization and invariant risk minimization. Sec.~\ref{sec:architecture overview} provides an overview of our architecture, followed by a detailed description of the mutual information mask module in Sec.~\ref{sec:mutual information mask}. Finally, we present details of the dual alignment optimization strategy for domain and modality in Sec.~\ref{sec:dual alignment of domain and modality}.

\subsection{Preliminaries}
\label{sec:preliminaries}
\textbf{Mutual Information Maximization.} The mutual information (MI) between two variables $X$ and $Y$ can be expressed as the difference between their entropy terms:
\begin{equation} 
\label{equ:mutual information 1}
\mathrm{I}(X;Y) = \mathrm{H}(X) - \mathrm{H}(X|Y) = \mathrm{H}(Y) - \mathrm{H}(Y|X)
\end{equation}
where $\mathrm{H}(\cdot)$ is the Shannon entropy, $\mathrm{H}(X|Y)$ denotes the conditional entropy of $X$ given $Y$. This definition has an intuitive interpretation: $\mathrm{I}(X;Y)$ is the reduction of uncertainty in $X$/$Y$ when $Y$/$X$ is observed. Alternatively, MI is also equivalent to the KL-divergence between the joint distribution $\mathrm{P}_{XY}$, and the product of the marginal distribution $\mathrm{P}_X\mathrm{P}_Y$:
\begin{equation} 
\label{equ:mutual information 2}
\mathrm{I}(X;Y) = \mathrm{D}_{\mathrm{KL}}(\mathrm{P}_{XY}||\mathrm{P}_X\mathrm{P}_Y)
\end{equation}
where the intuitive meaning of Eq.~\ref{equ:mutual information 2} is that the larger the divergence between the joint and the product of the marginals, the stronger the correlation between $X$ and $Y$.

For MI maximization, we typically need to estimate a lower bound of MI and then continuously raise this lower bound to achieve the goal. Belghazi et al. \cite{belghazi2018mutual} introduce a tighter MI Neural Estimator (MINE) based on Eq.~\ref{equ:mutual information 2} and Donsker-Varadhan representation theorem \cite{donsker1983asymptotic}, which converts them into the dual representation:
\begin{equation} 
\label{equ:Donsker-Varadhan}
\small
\mathrm{D}_{\mathrm{KL}}(\mathrm{P}||\mathrm{Q}) = \underset{T:\Omega \rightarrow \mathbb{R}}{\mathrm{sup}} \mathbb{E}_{\mathrm{P}}[T] - \mathrm{log}(\mathbb{E}_{\mathrm{Q}}[e^{T}]),
\end{equation}
\begin{equation} 
\label{equ:MINE}
\small
\mathrm{I}_{\mathrm{MINE}} = \mathbb{E}_{p(x,y)}[f(x,y)] - \mathrm{log}(\mathbb{E}_{p(x)p(y)}[e^{(f(x,y))}]),
\end{equation} 
where $\mathrm{P}$ and $\mathrm{Q}$ are two probability distributions, $T$ takes over all functions such that the two expectations are finite, and $f(\cdot,\cdot)$ represent a score function (or, critic) approximated by a neural network.

\noindent
\textbf{Empirical and Invariant Risk Minimization}. Empirical Risk Minimization (ERM) learning paradigm is widely used in machine learning, aiming to improve the model's generalization performance to unknown domains by minimizing the risk (i.e. loss) on the mixed source domain:
\begin{align}
\label{eqn:ERM}
\mathrm{ERM} \rightarrow \underset{\phi,\beta}{\mathrm{min}} \frac{1}{|\mathcal{E}|} \sum_{e \in \mathcal{E}} R^e(\phi,\beta),
\end{align}
where $\phi$ represents a neural network, $\beta$ denotes the hyperplane for classification, $\mathcal{E}$ represents the entire environment, $e$ is one of the sub-environments, and $f(x;\beta,\phi)$ is the function processing $x$ via $\phi,\beta$ and obtaining $y$. The empirical risk function $R^e(\phi,\beta)$, based on the loss function $\mathcal{L}(\cdot,\cdot)$, for a given environment $e$, is defined as:
\begin{align}
\label{eqn:ERM function}
R^e(\phi,\beta) = \mathbb{E}_{(x,y)\sim
e}[\mathcal{L}(f(x;\beta,\phi),y)].
\end{align}

While ERM is simple and effective, the i.i.d. assumption limits its application in FAS. It tends to fit all source training data together and undesirably leverages spurious correlations that may lead to poor generalization when test environments diverge from training data (i.e., domain shifts).

However, Invariant Risk Minimization (IRM) \cite{ahuja2021invariance,arjovsky2019invariant,choe2020empirical,krueger2021out,mitrovic2020representation} have been proposed to learn both unified and aligned classification hyperplane that is globally (for mixed all domains) and also locally (for each sub-domain) optimal. Specifically, the objective of IRM can be formulated as the following constrained optimization problem:
\begin{align}
\label{eqn:IRM}
\mathrm{IRM} \rightarrow \underset{\phi,\beta^*}{\mathrm{min}} \frac{1}{|\mathcal{E}|} \sum_{e \in \mathcal{E}} R^e(\phi,\beta^*), \notag \\
s.t. \, \beta^* \in \underset{\beta}{\mathrm{argmin}}R^e(\phi,\beta), \, \forall e \in \mathcal{E}.
\end{align}

Indeed, IRM is a challenging, bi-level optimization problem, which is hard to solve \cite{kamath2021does,rosenfeld2020risks}. Sun et. al \cite{sun2023rethinking} propose an equivalent objective Projected Gradient Optimization \cite{nocedal1999numerical} for IRM (PG-IRM) which is easier to optimize and achieve strong performance.

\subsection{Architecture Overview}
\label{sec:architecture overview}
Fig.~\ref{fig:architecture} shows our architecture. Our model builds upon ViT \cite{dosovitskiy2020image} by utilizing frozen pre-trained weights and introducing adapters for fine-tuning. Specifically, each layer of ViT comprises Layer Normalization (LN), Multi-Head Self Attention (MHSA), and Multi-Layer Perceptron (MLP). The model handles three input modalities: RGB, depth, and infrared images. Extracting the feature of each modality from the MHSA layer and feeding it into the Mutual Information Mask (MIM) module. Each MIM module receives two modalities, as illustrated in Fig.~\ref{fig:architecture}~(b). To ensure cross-modal fusion among all modalities, each layer includes three MIM modules. Fig.~\ref{fig:architecture}~(a) omits the schematic diagram of the interaction between infrared and depth, and the full interaction relationship is illustrated in Fig.~\ref{fig:mim}. In Fig.~\ref{fig:architecture}~(c), the CDC (short for Central Difference Convolution)-Adapter \cite{cai2023rehearsal} consists of vanilla and central differential convolution layers. In FAS tasks, the effectiveness of central differential convolution \cite{yu2020searching,yu2020fas,cai2023rehearsal} over vanilla ones have been well-verified, as it captures both intensity-level information and gradient-level messages, which are critical for distinguishing between live and spoof traces.

\subsection{Mutual Information Mask}
\label{sec:mutual information mask}
As mentioned in Sec.~\ref{sec:introduction}, the reliability of each modality can fluctuate based on the type of attack, rigid or uniform treatment of each modality is undesirable. Hence, the model should adaptively prioritize specific modalities or regions based on their reliability. To achieve this, we propose the Mutual Information Mask (MIM) module, which dynamically emphasizes reliable modalities and suppresses unreliable ones by leveraging mutual information maximization. 

Specifically, given a group of RGB, depth (D), infrared (I) images $\textbf{x}_{\mathrm{RGB}} \in \mathbb{R}^{H \times W \times 3}, \textbf{x}_{\mathrm{D}} \in \mathbb{R}^{H \times W \times 3}, \textbf{x}_{\mathrm{I}} \in \mathbb{R}^{H \times W \times 3}$, we split them into $\frac{H}{P} \times \frac{W}{P}$ non-overlapping patches $[\textbf{x}_{\mathrm{RGB}}^i]_{i=1}^{hw},[\textbf{x}_\mathrm{D}^i]_{i=1}^{hw},[\textbf{x}_\mathrm{I}^i]_{i=1}^{hw}$, where $H$, $W$ are the height and width, $P$ is the patch size, $h=\frac{H}{P}, w=\frac{W}{P}$. These patches are linearly projected to embedding vectors $[\textbf{z}_{\mathrm{RGB}}^i]_{i=1}^{hw} \in \mathcal{R}^{d},[\textbf{z}_\mathrm{D}^i]_{i=1}^{hw} \in \mathcal{R}^{d},[\textbf{z}_\mathrm{I}^i]_{i=1}^{hw} \in \mathcal{R}^{d}$:
\begin{equation}
\label{eqn: patch embedding}
[\textbf{z}_{\mathrm{RGB}}^i, \textbf{z}_{\mathrm{D}}^i, \textbf{x}_{\mathrm{I}}^i] = \mathrm{PE}([\textbf{x}_{\mathrm{RGB}}^i, \textbf{z}_{\mathrm{D}}^i, \textbf{x}_{\mathrm{I}}^i]) + \textbf{e}^i_{\mathrm{pos}},
\end{equation}
where $\mathrm{PE}(\cdot)$ is the patch embedding, and $\textbf{e}^i_{\mathrm{pos}} \in \mathbb{R}^{d}$ is the positional embedding. Then, we also introduce class tokens for all modalities $\textbf{z}_{\mathrm{RGB}}^0, \textbf{z}_\mathrm{D}^0, \textbf{z}_\mathrm{I}^0$.

\begin{figure}[t]
\vspace{-0.8em}
    \includegraphics[width=0.48\textwidth]{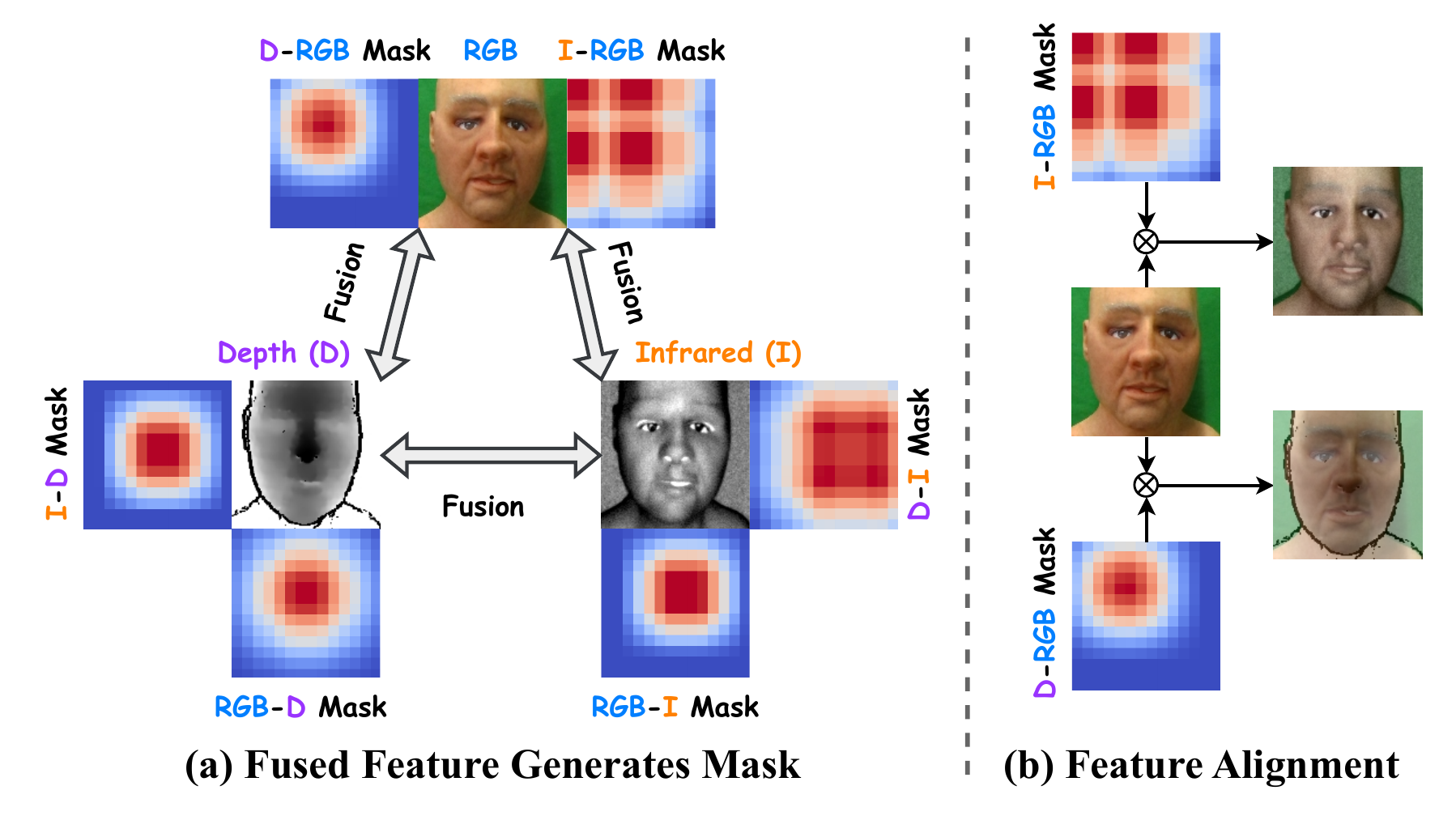}
    \vspace{-2.2em}
    \centering
    \caption{Alignment: (a) Mask generation via modality interaction. (b) Feature alignment via reweighting.}
    \label{fig:mim}
    \vspace{-0.3cm}
\end{figure}

The structure of MIM module is illustrated in Fig.~\ref{fig:architecture}~(b). We feed the outputs of each MHSA layer to MIM module for intra-domain modality alignment. At each MIM module, we first obtain the interactive feature of each modality pair via directly concatenating them along the channel dimension and feed into a lightweight interactive convolution block. Then, the interactive feature is fed into mask generation block $\mathrm{MG}(\cdot)$. The output is two informative masks corresponding to the modality pair:
\begin{align} 
\label{equ:informative point}
\textbf{z}_{\mathrm{fused}} = \mathrm{Conv}(\mathrm{Cat}(\textbf{z}_{m1},\textbf{z}_{m2})), \\
[\textbf{m}_{m1},\textbf{m}_{m2}] = \mathrm{sigmoid}(\mathrm{MG}(\textbf{z}_{\mathrm{fused}}))
\end{align}
where the subscript $m1$ and $m2$ represent two modalities, respectively. $\mathrm{Conv}(\cdot)$ and $\mathrm{MG}(\cdot)$ consists of a sequence of convolution and normalization layers. $\mathrm{sigmoid}(\cdot)$ is the activation function. We consider that the region with a higher weight (informative point), the more reliable information it involves. Conversely, regions associated with lower weights may carry redundant or negative information.

Then, we employ two masks $\textbf{m}_{m1},\textbf{m}_{m2} \in \mathbb{R}^{h \times w}$ to reweight the original features: 
\begin{equation} 
\label{equ:aligned feature}
\textbf{z}_{\mathrm{aligned}\_ m1} = \textbf{m}_{m1}\textbf{z}_{m1},\textbf{z}_{\mathrm{aligned}\_m2} = \textbf{m}_{m2}\textbf{z}_{m2}.
\end{equation}
Fig.~\ref{fig:mim} illustrates the details of our mutual information mask process. Then, the $\textbf{z}_{\mathrm{aligned}\_ m1},\textbf{z}_{\mathrm{aligned}\_ m2}$ are averaged as the mutual information (MI) tokens:
\begin{align} 
\label{equ:mutual information token}
z_\mathrm{mi1} = \bar{\textbf{z}}_{\mathrm{aligned}\_ m1},z_\mathrm{mi2} = \bar{\textbf{z}}_{\mathrm{aligned}\_ m2}.
\end{align}
Finally, we integrate interactive features of each modality pair with aligned features as the output of MIM module:
\begin{align} 
\label{equ:out feature}
\textbf{z}_{\mathrm{out}\_ m1} = \mathrm{Conv}(\mathrm{Cat}(\textbf{z}_{\mathrm{fused}},\textbf{z}_{\mathrm{aligned}\_ m1})),\\
\textbf{z}_{\mathrm{out}\_ m2} = \mathrm{Conv}(\mathrm{Cat}(\textbf{z}_{\mathrm{fused}},\textbf{z}_{\mathrm{aligned}\_ m2})).
\end{align}
We further alleviate intra-domain modality misalignment by MI maximization between MI tokens, as depicted in Eq.~\ref{equ:mutual information token}. The commonly used MINE \cite{belghazi2018mutual} needs a task-independent neural network to estimate the score function $f(\cdot,\cdot)$ of the joint and marginal distribution of $x,y$, which may be an unnecessary burden for the overall model. Coincidentally, we have obtained the MI tokens $z_{\mathrm{mi1}},z_{\mathrm{mi2}}$ in Eq.~\ref{equ:mutual information token}. We estimate the distribution entropy by taking the average of MI tokens, which can be seen as a special case of the score function. Thus we can rewrite Eq.~\ref{equ:MINE}:
\begin{align} 
\label{equ:MINE loss}
\mathcal{L}_{\mathrm{mi}} = -[\mathbb{E}_{p(z_\mathrm{mi1},z_\mathrm{mi2})}[\frac{z_\mathrm{mi1}+z_\mathrm{mi2}}{2}] \notag \\ - \mathrm{log}(\mathbb{E}_{p(z_\mathrm{mi1})p(z_\mathrm{mi2})}[e^{\frac{z_\mathrm{mi1}+z_\mathrm{mi2}}{2}}])].
\end{align}

During forward propagation, we calculate the $\mathcal{L}^l_{\mathrm{mi}}$ of each layer, and then average $\mathcal{L}^l_{\mathrm{mi}}$ from all layers as the final $\mathcal{L}_{\mathrm{mi}}$. In addition, we employ a gradient modulation technique, ReGrad \cite{lin2024suppress}, to adapt the optimization direction of each MIM module. Unlike \cite{lin2024suppress}, our ReGrad is based on the intensity of MI tokens instead of uncertainty in modality:
\begin{align}
\small
& \mathrm{ReGrad}(\textbf{g}_{1}, \textbf{g}_{2}) = \notag \\
& \left\{
\begin{aligned} 
& \textbf{g}_{1} + \frac{\textbf{g}_1 \cdot \textbf{g}_2}{||\textbf{g}_1||^2_2} \textbf{g}_1 \cdot \mathrm{mi}_2, \, \text{if } \textbf{g}_1 \cdot \textbf{g}_2 < 0, \mathrm{mi}_1 < \mathrm{mi}_2\\ 
& \textbf{g}_1 + (\textbf{g}_2 - \frac{\textbf{g}_1 \cdot \textbf{g}_2}{||\textbf{g}_1||^2_2} \textbf{g}_1) \cdot \mathrm{mi}_2,  \, \text{if } \textbf{g}_1 \cdot \textbf{g}_2 > 0, \mathrm{mi}_1 < \mathrm{mi}_2 \\ 
& \frac{\textbf{g}_1 \cdot \textbf{g}_2}{||\textbf{g}_2||^2_2} \textbf{g}_2  \cdot \mathrm{mi}_1  + \textbf{g}_2, \, \text{if } \textbf{g}_1 \cdot \textbf{g}_2 < 0, \mathrm{mi}_1 > \mathrm{mi}_2 \\ 
& (\textbf{g}_1 - \frac{\textbf{g}_1 \cdot \textbf{g}_2}{||\textbf{g}_2||^2_2} \textbf{g}_2) \cdot \mathrm{mi}_1  + \textbf{g}_2, \, \text{if } \textbf{g}_1 \cdot \textbf{g}_2 > 0, \mathrm{mi}_1 > \mathrm{mi}_2 \\ 
\end{aligned} 
\right.
\end{align}
where $\textbf{g}_1$ and $\textbf{g}_2$ denote gradient of each modality, respectively. $\mathrm{mi}_1=\bar{\textbf{z}}_{\mathrm{mi1}}$, $\mathrm{mi}_2=\bar{\textbf{z}}_{\mathrm{mi2}}$. The higher value of $\mathrm{mi}$, the more reliable its corresponding aligned features become, thus assigning greater significance to the gradient.

\subsection{Dual Alignment of Domain and Modality}
\label{sec:dual alignment of domain and modality}
Numerous uni-modal DG-based methods have been deeply explored to learn domain-invariant liveness representations over recent years. They typically adopt the Empirical Risk Minimization (ERM) learning paradigm. However, the i.i.d. assumption of ERM limits its application in the presence of significant domain shift tasks, e.g., FAS. Moreover, these methods commonly mix all source domains together and posit that the feature space becomes perfectly domain-invariant after removing the domain-specific signals. However, this approach has a significant drawback: when the source data is limited and target data exhibits high domain variability, performance can deteriorate substantially. This is because mixing source domains makes the feature space ambiguous, the live/spoof classifier may inadvertently rely on spurious correlations \cite{sun2023rethinking}, as shown in Fig.~\ref{fig:motivation} (a) and (c).

\begin{figure}[t!]
    \includegraphics[width=0.38\textwidth]{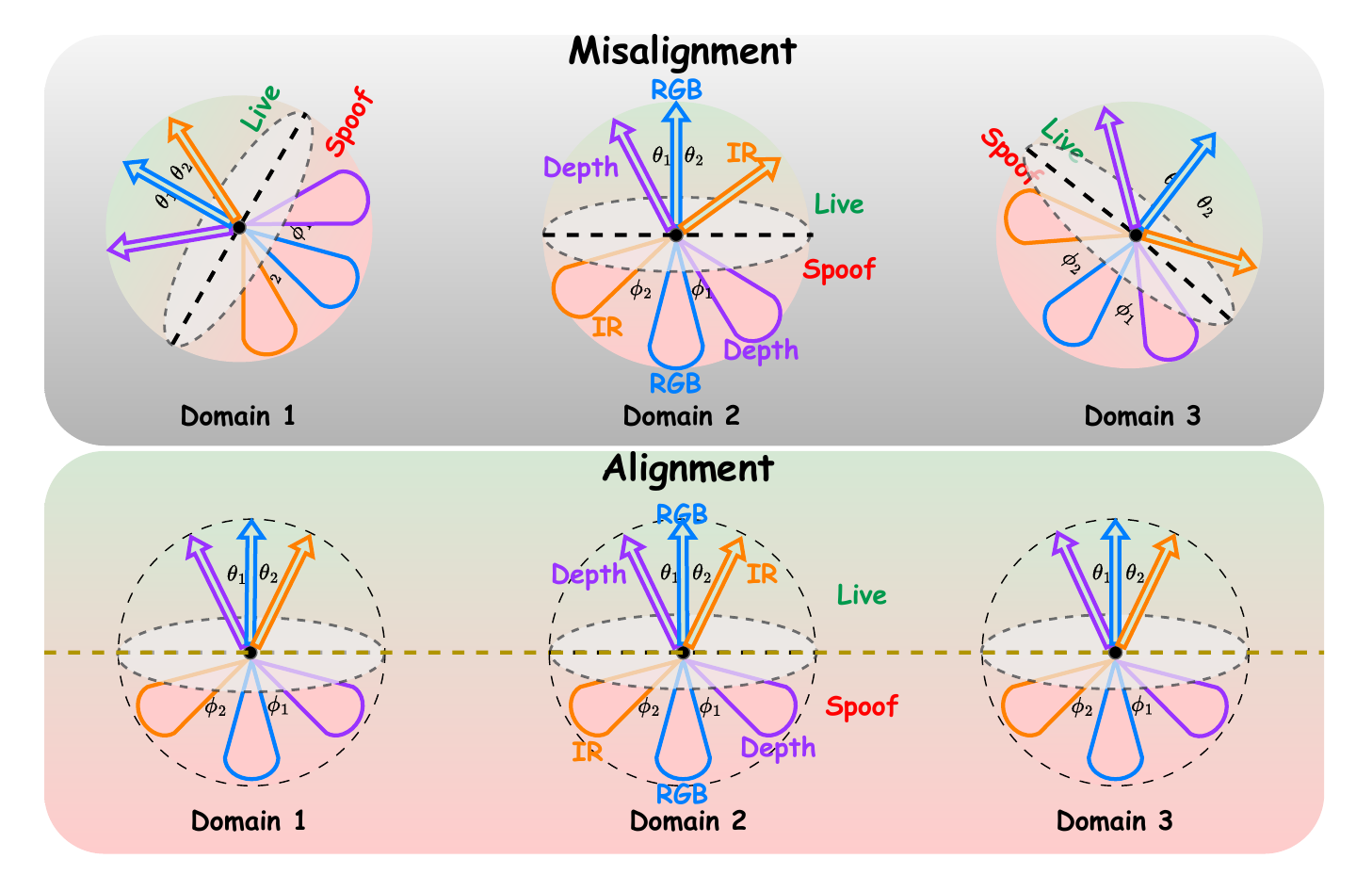}
    \vspace{-0.4cm}
    \centering
    \caption{Dual alignment of domain and modality. Samples from different domains are distributed on the same 3D sphere. We draw them separately to clearly indicate that varies domains have misaligned directions before alignment.}
    \label{fig:dadm}
    \vspace{-0.5cm}
\end{figure}

One competitive alternative is Invariant Risk Minimization (IRM), which aims to align the live-to-spoof transition to be the same for all domains, as illustrated in Fig.~\ref{fig:dadm}. We utilize the PG-IRM \cite{sun2023rethinking} algorithm to optimize the IRM problem, which constrains the alignment between the global domain optimal hyperplane and the local domain optimal hyperplane. For uni-modal FAS, alignment typically focuses only on the classification hyperplane. However, in multi-modal FAS, misalignments across domains become more pronounced. Significant domain shifts (angle deviations) in any modality can drastically affect overall performance. Hence, before obtaining the top-level features of each modality and fuse them for classification. We constrain the angle margins of modalities in each domain should be as consistent as possible, which is beneficial for the generalization. The angle margin loss is as follows:
\begin{align}
\label{eqn:angle}
\small
\mathcal{L}_{\mathrm{angle}} & = \sum_{e_1 \neq e_2}  \sum_{i \neq j}^{ \{\mathrm{RGB},\mathrm{D},\mathrm{I}\}} \mathbb{I}(y_i=1) \cdot (\frac{\textbf{z}^{e_1}_i \cdot \textbf{z}^{e_2}_i}{||\textbf{z}^{e_1}_i|| ||\textbf{z}^{e_2}_i||}-\tau_{\mathrm{l}})^2 \notag \\
&  + \mathbb{I}(y_j=0) \cdot (\frac{\textbf{z}^{e_1}_j \cdot \textbf{z}^{e_2}_j}{||\textbf{z}^{e_1}_j|| ||\textbf{z}^{e_2}_j||}-\tau_{\mathrm{s}})^2 + \notag \\
& \mathbb{I}(y_i=y_j) \cdot (\frac{\textbf{z}^{e_1}_i \cdot \textbf{z}^{e_1}_j}{||\textbf{z}^{e_1}_i|| ||\textbf{z}^{e_1}_j||} - \frac{\textbf{z}^{e_2}_i \cdot \textbf{z}^{e_2}_j}{||\textbf{z}^{e_2}_i|| ||\textbf{z}^{e_2}_j||})^2.
\end{align}
where $\mathbb{I}(y=1)$ denotes live conditions, $\mathbb{I}(y=0)$ denotes spoofs. The first term constrains all live samples to exhibit consistent angles, the second term constrains spoofs to exhibit relaxed consistent angles, and the third term constrains the angle difference among modalities of each domain is consistent. $\tau_l=1,\tau_s=0.85$ are used to control the relaxation degree of live and spoof samples, respectively.

\begin{table*}[t!]
\centering
\caption{Cross-dataset testing results under the fixed-modal scenarios (\textbf{Protocol 1}) among CASIA-CeFA (\textbf{C}), PADISI (\textbf{P}), CASIA-SURF (\textbf{S}), and WMCA (\textbf{W}). DG, MM, and FM are short for domain-generalized, multi-modal, and flexible-modal, respectively.}
\vspace{-0.2cm}
\label{table: protocol 1}
\setlength{\arrayrulewidth}{1.5pt}
\resizebox{0.96\textwidth}{!}{
\begin{tabular}{cccccccccccc}
\hline
\multirow{2}{*}{\textbf{Method}} & \multirow{2}{*}{\textbf{Type}} & \multicolumn{2}{c}{\textbf{CPS $\rightarrow$ W}} & \multicolumn{2}{c}{\textbf{CPW $\rightarrow$ S}} & \multicolumn{2}{c}{\textbf{CSW $\rightarrow$ P}} & \multicolumn{2}{c}{\textbf{PSW $\rightarrow$ C}} & \multicolumn{2}{c}{\textbf{Average}} \\
\cmidrule(r){3-4} \cmidrule(r){5-6} \cmidrule(r){7-8} \cmidrule(r){9-10} \cmidrule(r){11-12}
&  & HTER(\%) $\downarrow$ & AUC(\%) $\uparrow$ & HTER(\%) $\downarrow$ & AUC(\%) $\uparrow$ & HTER(\%) $\downarrow$ & AUC(\%) $\uparrow$ & HTER(\%) $\downarrow$ & AUC(\%) $\uparrow$ & HTER(\%) $\downarrow$ & AUC(\%) $\uparrow$ \\
\cmidrule(r){1-12}
SSDG~\cite{jia2020single} & DG  & 26.09  & 82.03  & 28.50  & 75.91 & 41.82 & 60.56 & 40.48  & 62.31 & 34.22 & 70.20 \\
SSAN~\cite{wang2022domain} & DG  & 17.73 & 91.69  & 27.94  & 79.04 & 34.49  & 68.85 & 36.43  & 69.29 & 29.15 & 77.22 \\
SA-FAS~\cite{sun2023rethinking} & DG  & 21.37 & 87.65 & 23.22& 84.49 & 35.10 & 70.86 & 35.38 & 69.71 & 28.77 & 78.18 \\
IADG~\cite{zhou2023instance} & DG  & 27.02  & 86.50 & 23.04 & 83.11 & 32.06 & 73.83 & 39.24 & 63.68 & 30.34 & 76.78 \\
ViTAF~\cite{huang2022adaptive} & DG  & 20.58  & 85.82  & 29.16  & 77.80 & 30.75 & 73.03 & 39.75  & 63.44 & 30.06 & 75.02 \\
CLIP~\cite{radford2021learning} & MM & 14.55 & 90.47 & 18.17 & 90.02 & 24.13 & 83.15 & 38.33  & 65.71 & 23.80 & 82.33 \\
FLIP~\cite{srivatsan2023flip} & MM & 13.19 & 93.79 & 11.73 & 94.93 & \textbf{17.39} & \textbf{90.63} & 22.14  & 83.95 & 16.11 & 90.83  \\
MM-CDCN~\cite{yu2020multi} & MM  & 38.92 & 65.39 & 42.93 & 59.79 & 41.38 & 61.51 & 48.14 & 53.71 & 42.84 & 60.10 \\
CMFL~\cite{george2021cross} & MM  & 18.22 & 88.82 & 31.20 & 75.66 & 26.68 & 80.85 & 36.93  & 66.82 & 28.26 & 78.04 \\
ViT+AMA~\cite{yu2024rethinking} & FM  & 17.56 & 88.74 & 27.50 & 80.00 & 21.18 & 85.51 & 47.48  & 55.56 & 28.43 & 77.45 \\
VP-FAS~\cite{yu2023visual} & FM & 16.26 & 91.22 & 24.42 & 81.07 & 21.76 & 85.46 & 39.35  & 66.55 & 25.45 & 81.08 \\
MMDG~\cite{lin2024suppress} & MM-DG  & 12.79 & 93.83 & 15.32 & 92.86 & 18.95 & 88.64 & 29.93  & 76.52 & 19.25 & 87.96  \\
\cmidrule(r){1-12}
\textbf{DADM (Ours)}& MM-DG & \textbf{11.71}& \textbf{94.89}  & \textbf{6.92}  & \textbf{97.66} & 19.03  & 88.22 & \textbf{16.87} & \textbf{91.08} & \textbf{13.63}& \textbf{92.96} \\
\hline
\end{tabular}
}
\vspace{-0.2cm}
\end{table*}

\begin{table*}[t!]
\centering
\caption{Cross-dataset testing results under the missing-modal scenarios (\textbf{Protocol 2}) among CASIA-CeFA (\textbf{C}), PADISI (\textbf{P}), CASIA-SURF (\textbf{S}), and WMCA (\textbf{W}). We report the average HTER (\%) and AUC(\%) on four sub-protocols, i.e. \textbf{CPS $\rightarrow$ W}, \textbf{CPW $\rightarrow$ S}, \textbf{CSW $\rightarrow$ P}, \textbf{PSW $\rightarrow$ C}. DG, MM, and FM are short for domain-generalized, multi-modal, and flexible-modal, respectively.}
\vspace{-0.2cm}
\label{table: protocol 2}
\setlength{\arrayrulewidth}{1.5pt}
\resizebox{0.84\textwidth}{!}{
\begin{tabular}{cccccccccccc}
\hline
\multicolumn{2}{c}{\multirow{2}{*}{\textbf{Method}}} & \multicolumn{2}{c}{\multirow{2}{*}{\textbf{Type}}} & \multicolumn{2}{c}{\textbf{Missing D}} & \multicolumn{2}{c}{\textbf{Missing I}} & \multicolumn{2}{c}{\textbf{Missing D \& I}} & \multicolumn{2}{c}{\textbf{Average}} \\
\cmidrule(r){5-6} \cmidrule(r){7-8} \cmidrule(r){9-10} \cmidrule(r){11-12}
& & & & HTER(\%) $\downarrow$ & AUC(\%) $\uparrow$ & HTER(\%) $\downarrow$ & AUC(\%) $\uparrow$ & HTER(\%) $\downarrow$ & AUC(\%) $\uparrow$ & HTER(\%) $\downarrow$ & AUC(\%) $\uparrow$ \\
\cmidrule(r){1-12}
\multicolumn{2}{c}{SSDG~\cite{jia2020single}} & \multicolumn{2}{c}{DG}  & 38.92  & 65.45  & 37.64  & 66.57 & 39.18 & 65.22 & 38.58  & 65.75 \\
\multicolumn{2}{c}{SSAN~\cite{wang2022domain}} & \multicolumn{2}{c}{DG} & 36.77 & 69.21  & 41.20 & 61.92 & 33.52 & 73.38 & 37.16 & 68.17 \\
\multicolumn{2}{c}{SA-FAS~\cite{sun2023rethinking}} & \multicolumn{2}{c}{DG} & 36.30 & 69.07 & 39.80 & 62.69 & 33.08 & 74.29 & 36.40 & 68.68 \\
\multicolumn{2}{c}{IADG~\cite{zhou2023instance}} & \multicolumn{2}{c}{DG}  & 40.72 & 58.72 & 42.17 & 61.83 & 37.50 & 66.90 & 40.13 & 62.49 \\
\multicolumn{2}{c}{ViTAF~\cite{huang2022adaptive}} & \multicolumn{2}{c}{DG}  & 34.99 & 73.22 & 35.88 & 69.40 & 35.89 & 69.61 & 35.59 & 70.64 \\
\multicolumn{2}{c}{MM-CDCN~\cite{yu2020multi}} & \multicolumn{2}{c}{MM}  & 44.90 & 55.35 & 43.60 & 58.38 & 44.54 & 55.08 & 44.35 & 56.27 \\
\multicolumn{2}{c}{CMFL~\cite{george2021cross}} & \multicolumn{2}{c}{MM}  & 31.37 & 74.62 & 30.55 & 75.42 & 31.89 & 74.29 & 31.27 & 74.78 \\
\multicolumn{2}{c}{ViT+AMA~\cite{yu2024rethinking}} & \multicolumn{2}{c}{FM}  & 29.25 & 77.70 & 32.30 & 74.06 & 31.48 & 75.82 & 31.01 & 75.86 \\
\multicolumn{2}{c}{VP-FAS~\cite{yu2023visual}} & \multicolumn{2}{c}{FM}  & 29.13 & 78.27 & 29.63 & 77.51 & 30.47 & 76.31 & 29.74 & 77.36 \\
\multicolumn{2}{c}{MMDG~\cite{lin2024suppress}} & \multicolumn{2}{c}{MM-DG}  & 24.89 & 82.39 & 23.39 & 83.82 & 25.26 & 81.86 & 24.51 & 82.69  \\
\cmidrule(r){1-12}
\multicolumn{2}{c}{\textbf{DADM (Ours)}} & \multicolumn{2}{c}{MM-DG} & \textbf{21.56}  & \textbf{85.17} & \textbf{20.82} & \textbf{85.28} & \textbf{22.61} & \textbf{84.04} & \textbf{21.66} & \textbf{84.83} \\
\hline
\end{tabular}
}
\vspace{-1.2em}
\end{table*}

Fig.~\ref{fig:dadm} shows the dual alignment of domain and modality. Before alignment, the classification hyperplane and modal angle margin of each domain are anisotropic. After alignment, the unified classification hyperplane that is globally and also locally optimal, and the modal angles exhibit consistency. In supplementary materials, we elaborate on the importance of dual alignment. Please refer to Sec. \ref{sec:proofs of the necessity of domain alignment and angle alignment} for detailed proof.

\subsection{Training and Inference}
\label{sec:training and inference}
The overall losses can be written as: 
\begin{equation}
\label{eqn:training}
\mathcal{L}_{\mathrm{total}} = \mathcal{L}_{ce} + \lambda_{\mathrm{mi}} \mathcal{L}_{\mathrm{mi}} + \lambda_{\mathrm{angle}} \mathcal{L}_{\mathrm{angle}},
\end{equation}
where $\lambda_{\mathrm{mi}},\lambda_{\mathrm{angle}}$ are the coefficients of losses. We use PG-IRM~\cite{sun2023rethinking} to optimize the total loss. The detailed optimization pipeline is provided in \textit{supplementary material}.

At the inference stage, we use the mean hyperplane from $\beta_{e_1},\beta_{e_2},\cdots,\beta_{\mathcal{E}}$ to get the final score. Specifically, the output is given by:
\begin{equation}
\label{eqn:inference}
Score = \frac{1}{|\mathcal{E}|} \sum_{e \in \mathcal{E}} {\beta_e}^T\phi(\textbf{x}_{\mathrm{RGB}},\textbf{x}_{\mathrm{D}},\textbf{x}_{\mathrm{I}}).
\end{equation}


\section{Experiments}
\label{sec:experiments}
\subsection{Datasets, Protocols, and Performance Metrics}
\label{sec:datasets, protocols, and performance metrics}
We use four commonly used multi-modal datasets: CASIA-CeFA (C) \cite{liu2021casia}, PADISI-USC (P) \cite{rostami2021detection}, CASIA-SURF (S) \cite{zhang2020casia}, and WMCA (W) \cite{george2019biometric} to evaluate the DG performance. Each dataset comprises three modalities: RGB, depth (D), and infrared (I). We employ four protocols, i.e., fixed modalities, missing modalities, flexible modalities, and limited source domains. In \textbf{Protocol 1}, we use four multi-modal leave-one-out (LOO) sub-protocols across C, P, S, and W, following \cite{lin2024suppress}. For \textbf{Protocol 2}, we evaluate three test-time missing-modal scenarios, i.e., D is missing, I is missing, and both D and I are missing. \textbf{Protocol 3} extends this setup to consider three flexible scenarios where modalities are optionally missing during both training and testing. The probability of modality missing is 0.3 during training and testing. Finally, \textbf{Protocol 4} limits the number of source domains. The performance metrics are Half Total Error Rate (HTER) and Area Under the Curve (AUC).

\begin{table*}[t!]
\centering
\caption{Cross-dataset testing results under the flexible-modal scenarios (\textbf{Protocol 3}) among CASIA-CeFA (\textbf{C}), PADISI (\textbf{P}), CASIA-SURF (\textbf{S}), and WMCA (\textbf{W}). We report the average HTER (\%) and AUC (\%) on four sub-protocols, i.e. \textbf{CPS $\rightarrow$ W}, \textbf{CPW $\rightarrow$ S}, \textbf{CSW $\rightarrow$ P}, \textbf{PSW $\rightarrow$ C}. DG, MM, and FM are short for domain-generalized, multi-modal, and flexible-modal, respectively.}
\vspace{-0.2cm}
\label{table: protocol 3}
\setlength{\arrayrulewidth}{1.5pt}
\resizebox{0.84\textwidth}{!}{
\begin{tabular}{cccccccccccc}
\hline
\multicolumn{2}{c}{\multirow{2}{*}{\textbf{Method}}} & \multicolumn{2}{c}{\multirow{2}{*}{\textbf{Type}}} & \multicolumn{2}{c}{\textbf{Flexible D}} & \multicolumn{2}{c}{\textbf{Flexible I}} & \multicolumn{2}{c}{\textbf{Flexible D \& I}} & \multicolumn{2}{c}{\textbf{Average}} \\
\cmidrule(r){5-6} \cmidrule(r){7-8} \cmidrule(r){9-10} \cmidrule(r){11-12}
& & & & HTER(\%) $\downarrow$ & AUC(\%) $\uparrow$ & HTER(\%) $\downarrow$ & AUC(\%) $\uparrow$ & HTER(\%) $\downarrow$ & AUC(\%) $\uparrow$ & HTER(\%) $\downarrow$ & AUC(\%) $\uparrow$ \\
\cmidrule(r){1-12}
\multicolumn{2}{c}{SSDG~\cite{jia2020single}} & \multicolumn{2}{c}{DG}  & 34.79  & 68.23  & 33.64 & 69.40 & 35.02 & 68.19 & 34.48  & 68.61 \\
\multicolumn{2}{c}{SSAN~\cite{wang2022domain}} & \multicolumn{2}{c}{DG}  & 32.86 & 72.15  & 35.82 & 65.55 & 29.96 & 77.50 & 32.88 & 71.73 \\
\multicolumn{2}{c}{SA-FAS~\cite{sun2023rethinking}} & \multicolumn{2}{c}{DG}  & 32.43 & 73.05 & 35.35 & 67.46 & 29.57 & 78.28 & 32.45 & 72.93 \\
\multicolumn{2}{c}{IADG~\cite{zhou2023instance}} & \multicolumn{2}{c}{DG}  & 36.39 & 61.22 & 37.69 & 64.46 & 33.52 & 69.74 & 35.87 & 65.14 \\
\multicolumn{2}{c}{ViTAF~\cite{huang2022adaptive}} & \multicolumn{2}{c}{DG}  & 31.27 & 77.33  & 32.07 & 74.35 & 32.08 & 72.57 & 31.81  & 74.75 \\
\multicolumn{2}{c}{MM-CDCN~\cite{yu2020multi}} & \multicolumn{2}{c}{MM} & 40.13 & 60.93 & 40.96 & 60.86 & 43.81 & 57.42 & 41.63 & 59.74 \\
\multicolumn{2}{c}{CMFL~\cite{george2021cross}} & \multicolumn{2}{c}{MM}  & 30.04 & 74.79 & 29.31 & 75.63 & 30.50 & 75.45 & 29.95  & 75.29 \\
\multicolumn{2}{c}{ViT+AMA~\cite{yu2024rethinking}} & \multicolumn{2}{c}{FM}  & 28.43 & 78.01 & 30.09 & 75.12 & 29.95 & 76.62 & 29.49 & 76.58 \\
\multicolumn{2}{c}{VP-FAS~\cite{yu2023visual}} & \multicolumn{2}{c}{FM} & 27.54 & 80.19 & 28.28 & 80.50 & 28.04 & 78.36 & 27.95 & 79.68 \\
\multicolumn{2}{c}{MMDG~\cite{lin2024suppress}} & \multicolumn{2}{c}{MM-DG}  & 23.25 & 85.89 & 21.91 & 86.38 & 22.58 & 84.54 & 22.58 & 85.60  \\
\cmidrule(r){1-12}
\multicolumn{2}{c}{\textbf{DADM (Ours)}} & \multicolumn{2}{c}{MM-DG} & \textbf{19.27}  & \textbf{88.79} & \textbf{19.98} & \textbf{88.45} & \textbf{21.67} & \textbf{86.16} & \textbf{20.31} & \textbf{87.80} \\
\hline
\end{tabular}
}
\vspace{-0.2cm}
\end{table*}

\begin{table}[h!]
\vspace{-0.3cm}
\caption{Cross-dataset results under the limited source domain scenarios (\textbf{Protocol 4}) among CASIA-CeFA (\textbf{C}), PADISI-USC (\textbf{P}), CASIA-SURF (\textbf{S}), and WMCA (\textbf{W}).}
\label{table: protocol 4}
\vspace{-0.2cm}
\centering
\setlength{\arrayrulewidth}{1.8pt}
\resizebox{0.47\textwidth}{!}{
\begin{tabular}{ccccccc}
\hline
\multirow{2}{*}{\textbf{Method}} & \multirow{2}{*}{\textbf{Type}} & \multicolumn{2}{c}{\textbf{CW $\rightarrow$ PS}} & \multicolumn{2}{c}{\textbf{PS $\rightarrow$ CW}} \\
\cmidrule(r){3-4} \cmidrule(r){5-6}
& & HTER(\%) $\downarrow$ & AUC(\%) $\uparrow$ & HTER(\%) $\downarrow$ & AUC(\%) $\uparrow$ \\
\cmidrule(r){1-6}
SSDG~\cite{jia2020single} & DG & 25.34  & 80.17  & 46.98  & 54.29 \\
SSAN~\cite{wang2022domain} & DG & 26.55  & 80.06  & 39.10  & 67.19 \\
SA-FAS~\cite{sun2023rethinking} & DG & 25.20 & 81.06 & 36.59 & 70.03 \\
IADG~\cite{zhou2023instance} & DG & 22.82  & 83.85  & 39.70  & 63.46 \\
ViTAF~\cite{huang2022adaptive} & DG  & 29.64  & 77.36  & 39.93  & 61.31 \\
MM-CDCN~\cite{yu2020multi} & MM  & 29.28 & 76.88 & 47.00 & 51.94 \\
CMFL~\cite{george2021cross} & MM  & 31.86 & 72.75 & 39.43 & 63.17 \\
ViT+AMA~\cite{yu2024rethinking} & FM  & 29.25 & 76.89 & 38.06 & 67.64 \\
VP-FAS~\cite{yu2023visual} & FM & 25.90 & 81.79 & 44.37 & 60.83  \\
MMDG~\cite{lin2024suppress} & MM-DG  & 20.12 & 88.24 & 36.60 & 70.35 \\
\cmidrule(r){1-6}
\textbf{DADM (Ours)} & MM-DG & \textbf{12.61}  & \textbf{93.81}  & \textbf{20.40}  & \textbf{89.51} \\
\hline
\end{tabular}
\vspace{-0.9cm}
}
\end{table}

\subsection{Implementation Details}
\label{sec:implementation details}
All RGB, depth, and infrared images are resized to 224 $\times$ 224 $\times$ 3. We employ ViT-B/16 as the backbone, pretrained on ImageNet with a patch size of 16 and a hidden size of 768. The model is trained for 50 epochs using the Adam optimizer \cite{kingma2014adam} with a learning rate of $5\times10^{-5}$ and a weight decay of $1\times10^{-3}$. The batch size is set to 32. The hyper-parameters involved are as follows: $\tau_l$=1.0, $\tau_s$=0.85, $\lambda_{\mathrm{mi}}$=0.1, and $\lambda_{\mathrm{angle}}$=0.3.

\subsection{Multi-Modal Cross-Domain Evaluation}
\label{sec:multi-modal cross-domain evaluation}
We compare our method on Protocols 1-4 against three categories of FAS methods: (a) Uni-modal domain generalization (DG) FAS.~\cite{jia2020single,wang2022domain,zhou2023instance,zhou2022adaptive} (b) Multi-modal FAS~\cite{george2021cross,yu2024rethinking}. (c) Common multi-modal domain generalization FAS, i.e., MMDG \cite{lin2024suppress}. For (a), to make them compatible with multi-modal protocols, concatenating all three modalities along the channel dimension and introducing a trainable convolution layer to adapt to a 3-channel input.

\begin{table}[t!]
\vspace{-0.2cm}
\caption{Ablation results on our proposed components. We report the average HTER and AUC.}
\label{table: ablation study 1}
\vspace{-0.3cm}
\centering
\setlength{\arrayrulewidth}{1.8pt}
\resizebox{0.49\textwidth}{!}{
\begin{tabular}{cccccc}
\hline
\textbf{Backbone} & \textbf{Adapter} & \textbf{ReGrad}& \textbf{Loss} & \textbf{HTER} (\%) $\downarrow$ & \textbf{AUC} (\%) $\uparrow$  \\
\hline
ViT & - & - & CE & 31.14 & 74.81 \\
ViT & U-Adapter~\cite{lin2024suppress} & - & CE+SSP~\cite{lin2024suppress} & 24.54& 83.14\\
ViT & MIM+Vanilla-Conv& - & CE+SSP~\cite{lin2024suppress} & 22.97& 85.29\\
ViT & MIM+CDC-Adapter & - & CE+SSP~\cite{lin2024suppress} & 21.75& 86.17\\
ViT & U-Adapter & UEM-Guided & CE+SSP~\cite{lin2024suppress} & 19.25& 87.96\\
ViT & MIM+CDC-Adapter & MI-Guided & CE+SSP~\cite{lin2024suppress} +MI & 17.17& 88.14\\
ViT & MIM+CDC-Adapter & MI-Guided & PG-IRM~\cite{sun2023rethinking} & 16.54& 90.27\\
ViT & MIM+CDC-Adapter & MI-Guided & DADM & 14.31& 92.05\\
ViT & MIM+CDC-Adapter & MI-Guided & DADM+MI & \textbf{13.63}& \textbf{92.96}\\
\hline
\end{tabular}
}
\vspace{-0.4cm}
\end{table}

\noindent
\textbf{Protocol 1: Fixed-Modal Scenarios.} Tab.~\ref{table: protocol 1} shows our method achieves superior performance across sub-protocols. Compared to most uni-modal DG and multi-modal FAS methods, DADM demonstrates a significant improvement, with an average HTER improvement of 15.52\% and an average AUC improvement of 15.74\% compared to SSAN~\cite{wang2022domain}, and 11.82\% in HTER and 11.88\% in AUC over VP-FAS~\cite{yu2023visual}. Against previous state-of-the-art MMDG~\cite{lin2024suppress}, DADM consistently outperforms except on \textbf{CSW $\rightarrow$ P}, where the performance is slightly inferior. In particular, for \textbf{PSW $\rightarrow$ C}, our method achieves overwhelming advantages, with a 13.06\% HTER and 14.56\% AUC improvement. Overall, the average improvement across all sub-protocols is 5.62\% in HTER and 5\% in AUC.


\vspace{0.5mm}
\noindent
\textbf{Protocol 2: Missing-Modal Scenarios.} While recent multi-modal methods can improve the robustness of FAS systems, they require consistent training and testing modalities. This reliance overlooks scenarios where a modality may be missing during testing or in the real world, often leading to failure in distinguishing between live and spoof faces. In Protocol 2, when a modality is marked as missing, the input of that branch will be replaced with all zero input. As depicted in Table~\ref{table: protocol 2}, DADM outperforms all methods in scenarios where depth, infrared, or both of them are missing at test time. Notably, while ViT+AMA and VP-FAS are specifically designed for missing-modality settings, they do not consider the intra-domain modality and inter-domain modality misalignment and resulting in limited performance, while DADM handles effectively. 

\begin{table}[t]
    \vspace{-0.2cm}
    \begin{minipage}{0.235\textwidth}
        \centering
        \caption{Ablation results on different mutual information maximization losses.}
        \vspace{-0.2cm}
        \label{table: ablation study 2}
        \setlength{\arrayrulewidth}{1.2pt}
        \resizebox{0.99\textwidth}{!}{
        \begin{tabular}{ccc}
        \hline
        \textbf{Method} & \textbf{HTER} (\%) $\downarrow$ & \textbf{AUC} (\%) $\uparrow$  \\
        \hline
        InfoNCE \cite{oord2018representation}& 15.80& 91.47\\
        MINE \cite{belghazi2018mutual}& 14.40& 92.13\\
        \textbf{Ours} & \textbf{13.63}& \textbf{92.96}\\
        \hline
        \end{tabular}
        }
    \end{minipage}
    \hfill
    \begin{minipage}{0.235\textwidth}
        \centering
        \caption{Ablation results on vanilla and central difference convolutional adapter.}
        \vspace{-0.2cm}
        \label{table: ablation study 3}
        \setlength{\arrayrulewidth}{1.2pt}
        \resizebox{0.99\textwidth}{!}{
        \begin{tabular}{ccc}
        \hline
        \textbf{Method} & \textbf{HTER} (\%) $\downarrow$ & \textbf{AUC} (\%) $\uparrow$  \\
        \hline
        Vanilla Conv & 14.71& 91.63\\
        CDC Conv & \textbf{13.63}& \textbf{92.96}\\
        \hline
        \end{tabular}
        }
    \end{minipage}
\vspace{-0.4cm}
\end{table}

\begin{table}[t]
\caption{Ablation results on mask generation strategy.}
\label{table: ablation study 4}
\vspace{-0.3cm}
\centering
\setlength{\arrayrulewidth}{1.2pt}
\resizebox{0.49\textwidth}{!}{
\begin{tabular}{ccccc}
\hline
\textbf{Method} & Distance-based \cite{tu2023implicit} & Threshold-based & Attention-based \cite{lin2024suppress} & \textbf{Ours}\\
\hline
\textbf{HTER} (\%) $\downarrow$ & 14.08& 21.28& 15.74& \textbf{13.63}\\
\textbf{AUC} (\%) $\uparrow$ & 92.46& 83.19& 90.33& \textbf{92.96}\\
\hline
\end{tabular}

}
\vspace{-0.5cm}
\end{table}

\noindent
\textbf{Protocol 3: Flexible-Modal Scenarios.} The modality-missing in Protocol 2 only occurs during testing, while Protocol 3 may experience modality-missing during both training and testing. Additionally, we abandon the all-zero substitution approach in Protocol 2. Instead, when a modality is marked as missing during training, we leverage a learnable input tensor as a substitute, which has the same shape as the input images. This approach is somewhat similar to VP-FAS~\cite{yu2023visual}. As there are three modalities of input that may all be missing, we set three learnable tensors. Tab.~\ref{table: protocol 3} shows that DADM surpasses competing methods in scenarios with flexible-missing modalities. Our method attains an average HTER improvement of 9.18\% and an average AUC improvement of 11.22\% compared to ViT+AMA~\cite{yu2024rethinking}, and 7.64\% in HTER and 8.12\% in AUC over VP-FAS~\cite{yu2023visual}. Against MMDG~\cite{lin2024suppress}, our method achieves a 2.57\% HTER and 2.20\% AUC improvement.

\noindent
\textbf{Protocol 4: Limited Source Domains.} In Protocol 4, the limited source domains intensify domain shift severity. Our approach achieves optimal results across sub-protocols, as shown in Tab.~\ref{table: protocol 4}. Other methods exhibit significant performance degradation compared to Protocol 1, yet DADM maintains robust performance, especially under substantial domain shift in \textbf{PS $\rightarrow$ CW}. This further demonstrates the superior generalization of dual alignment of domain and modality, particularly in limited source data.

\begin{figure}[t!]
    \includegraphics[width=0.43\textwidth]{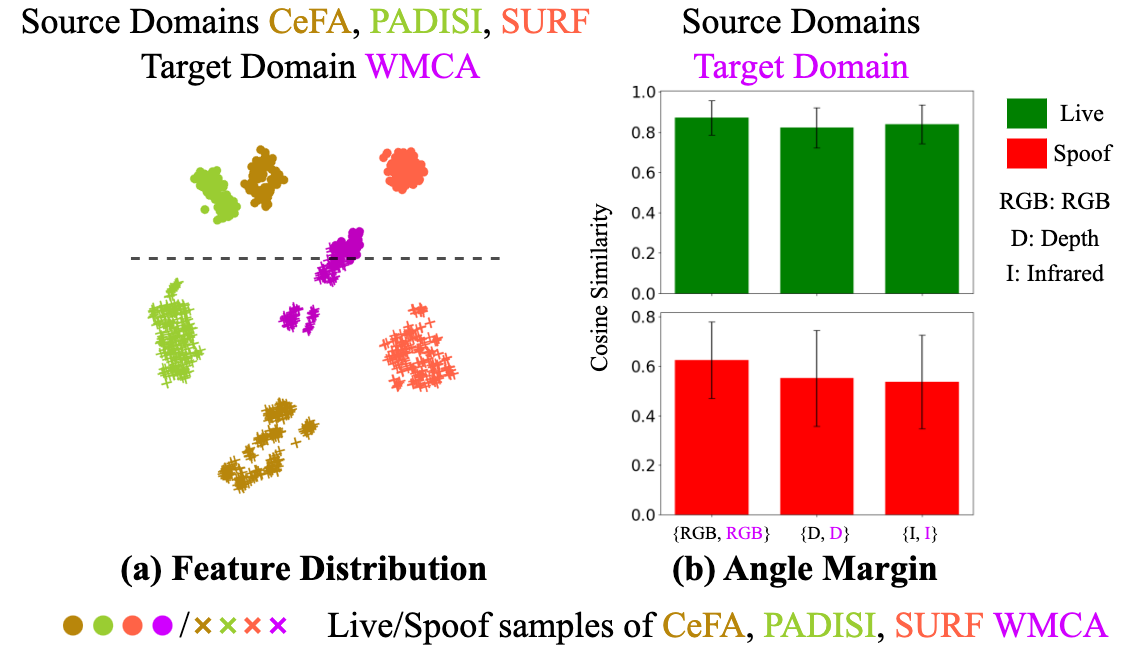}
    \vspace{-0.3cm}
    \centering
    \caption{(a) t-SNE result of feature distribution for source and target domains, the dotted line represents the decision hyperplane in 2D space. (b) Mean and Std. of cosine similarity.}
    \label{fig:tsne}
    \vspace{-0.6cm}
\end{figure}

\subsection{Ablation Study}
\label{sec:ablation study}
\textbf{Effectiveness of MIM, CDC-Adapter, and DADM optimization strategy}. Here, we perform a comprehensive ablation study on our proposed components MIM, CDC-Adapter, and DADM optimization strategy to demonstrate their individual effectiveness, as shown in Tab.~\ref{table: ablation study 1}. We utilize a vanilla ViT as the backbone, the model experiences a performance boost when additional U-Adapters \cite{lin2024suppress}, MIM+Vanilla-Conv-Adapter, or MIM+CDC-Adapter are employed. The MIM+CDC-Adapter achieves the best result among them. This demonstrates the effectiveness of MIM and CDC-Adapter \cite{cai2023rehearsal} for existing multi-modal DG FAS. After we employ MIM, the ReGrad \cite{lin2024suppress} switches to MI-guided mode, while U-Adapter \cite{lin2024suppress} enables UEM-guided \cite{lin2024suppress} mode. The performance improvement is further pronounced when our MI-guided ReGrad is integrated. As we gradually adopt the DADM optimization strategy, the points also steadily improve. Finally, by incorporating all the components, the performance reaches the optimal. 

\noindent
\textbf{Comparison of different mutual information losses and convolutional adapters}. Tab.~\ref{table: ablation study 2} compares some different mutual information maximization losses. Our experiments show that the InfoNCE \cite{oord2018representation}, which is based on contrastive learning MI maximization has limited performance, and the common MINE \cite{belghazi2018mutual} rely on the estimation of score function, introducing additional task-irrelevant network structures, does not achieve the best performance as well. However, the performance of our mentioned MI maximization loss in Eq.~\ref{equ:MINE loss} is optimal. In Tab.~\ref{table: ablation study 3}, we compare the Adapter based on vanilla convolution and CDC, and the results demonstrate that the performance of the CDC-Adapter is better. This is because CDC \cite{yu2020searching} combines both intensity-level semantic information and gradient-level messages, which are critical for capturing liveness representations. 

\noindent
\textbf{Performance of various mask generation strategies}. In Tab.~\ref{table: ablation study 4}, we attempt several other alignment mask strategies, such as attention-based \cite{lin2024suppress}, distance-based \cite{tu2023implicit}, and simple threshold-based. The results show that our mutual information-based alignment mask strategy is optimal for solving multi-modal DG-FAS tasks.

\begin{figure}[t!]
    \includegraphics[width=0.48\textwidth]{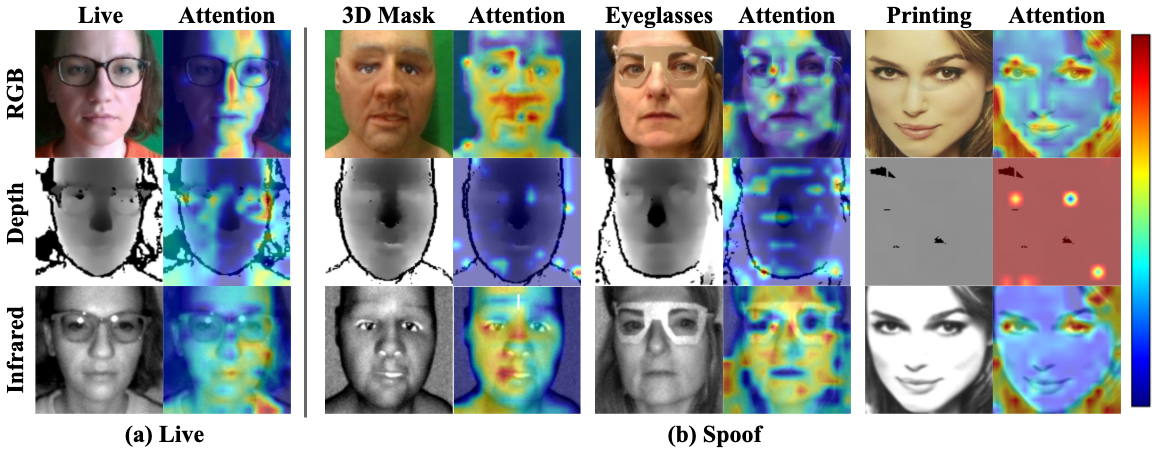}
    \vspace{-0.7cm}
    \centering
    \caption{Visual attention map of all modalities and different presentation types.}
    \label{fig:attention}
    \vspace{-0.5cm}
\end{figure}

\subsection{Visualization and Analysis}
\label{sec:visualization and analysis}
\textbf{Impact of dual alignment of domain and modality on feature distribution.} Fig.~\ref{fig:tsne} validates the dual alignment of classification hyperplane and angle margin among modalities. As Fig.~\ref{fig:tsne}~(a), we utilize t-SNE \cite{van2008visualizing} to perform feature visualization on \textbf{CPS $\rightarrow$ W}. The hyperplane between lives and spoofs is consistent across different source domains and also transferable to unknown domains. Moreover, the distribution of lives are more compact, while the spoofs are scattered. In Fig.~\ref{fig:tsne}~(b), the modality angles between source and target domains exhibit generalization. The vertical axis represents the cosine similarity, and the horizontal axis denotes the corresponding modality. The histogram depicts the mean and variance of cosine similarity (i.e., angle).

\vspace{0.5mm}
\noindent
\textbf{Visualization attention map demonstrates the function of the mutual information maximization.} We visualize attention maps to verify the mechanism of the MIM module by using Grad-CAM \cite{selvaraju2017grad}. As Fig.~\ref{fig:attention}~(b), for 3D mask attacks, the depth information of facial region is unreliable, thus it is assigned a lower importance. Similarly, for eyeglasses attacks, depth information is not reliable. Regarding printings attack, at the right of Fig.~\ref{fig:attention}~(b), depth is easy to distinguish spoof pattern, so it obtain the higher importance than others. Additionally, in Fig.~\ref{fig:attention} (a), for lives, three modalities are all useful, as they need to possess live features simultaneously to be judged as live. These observations validate the effectiveness of using mutual information masks can adaptively enhance reliable modalities and suppress unreliable ones. For More visualizations, please refer to supplementary materials.

\section{Conclusion}
\label{sec:conclusion}
In this paper, we propose a novel multi-modal DG-FAS framework to enhance generalization by addressing intra- and inter-domain misalignment of modalities. To mitigate the intra-domain modality misalignment, we propose mutual information mask to adaptively enhance favorable modalities and suppress unfavorable ones during cross-modal alignment. To mitigate the inter-domain modality misalignment, we dually align the hyperplane of each domain and the angle margin among modalities. Extensive experiments on multi-modal DG-FAS benchmark demonstrate the effectiveness of our method.

{
    \small
    \bibliographystyle{ieeenat_fullname}
    \bibliography{main}
}
\clearpage
\setcounter{page}{1}
\maketitlesupplementary

\section{Proofs of Donsker-Varadhan Representation Theorem}
\label{sec:proofs Donsker-Varadhan Representation Theorem}
We provide this section for helping understand mutual information maximization formula in Sec. \ref{sec:preliminaries}. We typically need to estimate a lower bound of mutual information and then continuously raise this lower bound to achieve the goal. Among them, Donsker-Varadhan theorem \cite{donsker1983asymptotic} is a commonly used estimation of the lower bound of mutual information. Belghazi et al. \cite{belghazi2018mutual} converts it into the dual representation:
\begin{equation} 
\label{equ:MINE_}
\mathrm{I}_{\mathrm{MINE}} = \mathbb{E}_{p(x,y)}[f(x,y)] - \mathrm{log}(\mathbb{E}_{p(x)p(y)}[e^{(f(x,y))}]).
\end{equation}

\noindent
\textbf{Donsker-Varadhan representation theorem} \cite{donsker1983asymptotic}. \textit{The KL-divergence possesses the following dual representation supremum:}
\begin{equation} 
\label{equ:Donsker-Varadhan representation}
\mathrm{D}_{\mathrm{KL}}(\mathrm{P}||\mathrm{Q}) = \underset{T:\, \Omega \rightarrow \mathbb{R}, \, T \in \mathcal{F}}{\mathrm{sup}} \mathbb{E}_{\mathrm{P}}[T] - \mathrm{log}(\mathbb{E}_{\mathrm{Q}}[e^{T}]),
\end{equation}
\textit{where the supremum is taken over all functions $T$ such that the two expectations ar finite. $\mathcal{F}$ be any class of functions $T:\Omega \rightarrow \mathbb{R}$ satisfying the integrability constrains of the theorem.}

For a given function $T$, consider the Gibbs distribution $\mathrm{G}$ defined by $\mathrm{dG}=\frac{1}{Z}e^T\mathrm{dQ}$, where $Z=\mathbb{E}_{\mathrm{Q}}[e^T]$. By construction
\begin{equation} 
\label{equ:Donsker-Varadhan representation 1}
\mathbb{E}_{\mathrm{P}}[T] - \mathrm{log}(\mathbb{E}_{\mathrm{Q}}[e^{T}]) = \mathbb{E}_{\mathrm{P}}[\mathrm{log}\frac{\mathrm{dG}}{\mathrm{dQ}}],
\end{equation}
as $T=\mathrm{log}[Z\frac{\mathrm{dG}}{\mathrm{dQ}}]=\mathrm{log}Z + \mathrm{log}\frac{\mathrm{dG}}{\mathrm{dQ}}=\mathrm{log}(\mathbb{E}_{\mathrm{Q}}[e^{T}])+ \mathrm{log}\frac{\mathrm{dG}}{\mathrm{dQ}}$. Let $\Delta$ be the gap, and combining Eqn. \ref{equ:Donsker-Varadhan representation 1}:
\begin{align} 
\label{equ:Donsker-Varadhan representation 2}
\Delta & = \mathrm{D}_{\mathrm{KL}}(\mathrm{P}||\mathrm{Q}) - \big(\mathbb{E}_{\mathrm{P}}[T] - \mathrm{log}(\mathbb{E}_{\mathrm{Q}}[e^{T}])\big), \\
\Delta & = \mathrm{D}_{\mathrm{KL}}(\mathrm{P}||\mathrm{Q}) - \mathbb{E}_{\mathrm{P}}[\mathrm{log}\frac{\mathrm{dG}}{\mathrm{dQ}}], \\
\Delta & = \mathbb{E}_{\mathrm{P}}[\mathrm{log}\frac{\mathrm{dP}}{\mathrm{dQ}} - \mathrm{log}\frac{\mathrm{dG}}{\mathrm{dQ}}] = \mathbb{E}_{\mathrm{P}}[\mathrm{log}\frac{\mathrm{dP}}{\mathrm{dG}}] = \mathrm{D}_{\mathrm{KL}}(\mathrm{P}||\mathrm{G}),
\end{align}
we can easily draw the conclusion that $\Delta \geq 0$, because KL-divergence $\mathrm{D}_{\mathrm{KL}}(\mathrm{P}||\mathrm{G})$ is always positive, i.e., $\mathrm{D}_{\mathrm{KL}}(\mathrm{P}||\mathrm{Q}) \geq \ \mathbb{E}_{\mathrm{P}}[T] - \mathrm{log}(\mathbb{E}_{\mathrm{Q}}[e^{T}])$. The proof is completed.

Since mutual information can be written in the form of the KL-divergence between the joint distribution and the product of the marginal distribution, such a lower bound can also be obtained for mutual information. The idea of \cite{belghazi2018mutual} is to choose $\mathcal{F}$ to be the family of functions parametrized by a deep neural network with parameters $\theta \in \Theta$, so there exists:
\begin{equation} 
\label{equ:mutual information relationship}
\mathrm{I}(X;Y) \geq \mathrm{I}_{\Theta}(X;Y),
\end{equation}
where $\mathrm{I}_{\Theta}(X;Y)$ is defined as:
\begin{equation} 
\label{equ:mutual information parametrized}
\mathrm{I}_{\Theta}(X;Y) = \underset{\theta \in \Theta}{\mathrm{sup}} \mathbb{E}_{\mathrm{P}_{XY}}[T_{\theta}] - \mathrm{log}(\mathbb{E}_{\mathrm{P}_X\mathrm{P}_Y}[e^{T_{\theta}}]).
\end{equation}
In code implementation, we estimate the expectations in Eq. \ref{equ:mutual information parametrized} using empirical samples from $\mathrm{P}_{XY}$ and $\mathrm{P}_X\mathrm{P}_Y$ (i.e., by shuffling the samples from the joint distribution along the batch axis). Ultimately, the objective function can be optimized through gradient descent and back propagation. The common approach is to use an independent neural network to process the features of two modalities $X$ and $Y$. Instead, we employ the average of the mutual information tokens mentioned in Eq. \ref{equ:mutual information token}, \ref{equ:MINE loss}, where the MI tokens represent the summarization of fused features. It can be seen as a special case of the score function $f(x,y)$.

\section{Supplementary Experimental Results}
\label{sec:supplementary experimental results}

\textbf{Empirical studies on hyper-parameters}. In Tab.~\ref{table: empirical study 1}, we conduct empirical studies on the $\lambda$ coefficients of different loss terms. We select appropriate $\lambda$ values within the interval of (0,1) to find the relatively optimal combination. The final combination obtained is $(\lambda_{\mathrm{mi}},\lambda_{\mathrm{angle}})=(0.1,0.3)$. In Tab.~\ref{table: empirical study 2}, we carry out empirical studies on temperature coefficients $\tau_l$ and $\tau_s$. We attempt various values for $\tau_l$ and $\tau_s$ within the interval of (0,1) to determine the relatively optimal combination. Meanwhile, based on the experience from some previous studies  \cite{yang2024generalized,jia2020single}, we assume that live and spoof samples exhibit an asymmetry distribution with different degree pf relaxation in the hyper-feature space. Therefore, when conducting our attempts, we prefer to impose a more compact feature distribution to the live samples, while allowing the spoof samples to have a looser feature distribution. The optimal combination we have found is $(\tau_l,\tau_s)=(1.0,0.85)$.

\begin{table}[h]
        \centering
        \caption{Empirical studies on $\lambda$ coefficients.}
        \vspace{-0.2cm}
        \label{table: empirical study 1}
        \setlength{\arrayrulewidth}{1.0pt}
        \resizebox{0.35\textwidth}{!}{
        \begin{tabular}{cc|cc}
        \hline
        $\lambda_{\mathrm{mi}}$ & $\lambda_{\mathrm{angle}}$ & HTER (\%) $\downarrow$ & AUC (\%) $\uparrow$  \\
        \hline
        0.3 & 0.5 & 15.54 & 90.54\\
        0.2 & 0.5 & 14.98 & 90.63\\
        0.1 & 0.5 & 14.33 & 91.65\\
        0.1 & 0.4 & 14.02 & 91.94\\
        0.0 & 0.4 & 14.52 & 92.11\\
        0.0 & 0.3 & 14.31 & 92.05\\
        0.1 & 0.3 & \textbf{13.63} & \textbf{92.96}\\
        \hline
        \end{tabular}
    }
\end{table}

 \begin{table}[h]
 \vspace{-0.2cm}
        \centering
        \caption{Empirical results on temperature coefficient $\tau_l$ and $\tau_s$.}
        \vspace{-0.2cm}
        \label{table: empirical study 2}
        \setlength{\arrayrulewidth}{1.0pt}
        \resizebox{0.35\textwidth}{!}{
        \begin{tabular}{cc|cc}
        \hline
        $\tau_l$ & $\tau_s$ & HTER (\%) $\downarrow$ & AUC (\%) $\uparrow$  \\
        \hline
        1.0 & 0.5 & 15.80& 90.77\\
        1.0 & 0.6 & 15.25& 90.68\\
        1.0 & 0.7 & 14.74& 91.05\\
        1.0 & 0.8 & 14.28& 91.93\\
        1.0 & 0.9 & 13.91& 92.30\\
        0.9 & 0.8 & 14.09& 91.92\\
        0.95 & 0.8 & 13.85& 91.98\\
        1.0 & 0.85 & \textbf{13.63}& \textbf{92.96}\\
        \hline
        \end{tabular}
        }
    \end{table}

\noindent
\textbf{Convergence speed of CDC-Adapter and vanilla convolutional Adapter}. 
In face anti-spoofing field, there are lots of works \cite{yu2020searching,yu2020multi,cai2023rehearsal}, apply central difference convolution operator for live/spoof representation capture. The CDC \cite{yu2020searching} operator combines both intensity-level semantic information and gradient-level messages:
\begin{align}
y(p_0) =  \, & \theta \cdot \underbrace{\sum_{p_n \in \mathcal{P}} w(p_n) \cdot (x(p_0+p_n)-x(p_0))}_{\text{central difference convolution}} \notag \\
& (1-\theta) \cdot \underbrace{\sum_{p_n \in \mathcal{P}} w(p_n) \cdot x(p_0+p_n)}_{\text{vanilla convolution}},
\end{align}
where $p_0$ is current location on input feature map while $p_n$ enumerates the locations in $\mathcal{P}$ (pixel neighborhood), $w(p_n)$ are the weights of convolutional kernel corresponding to the location $p_n$, hyper-parameter $\theta \in [0,1]$ tradeoffs the importance between intensity and gradient information.

\begin{figure}[t!]
    \includegraphics[width=0.49\textwidth]{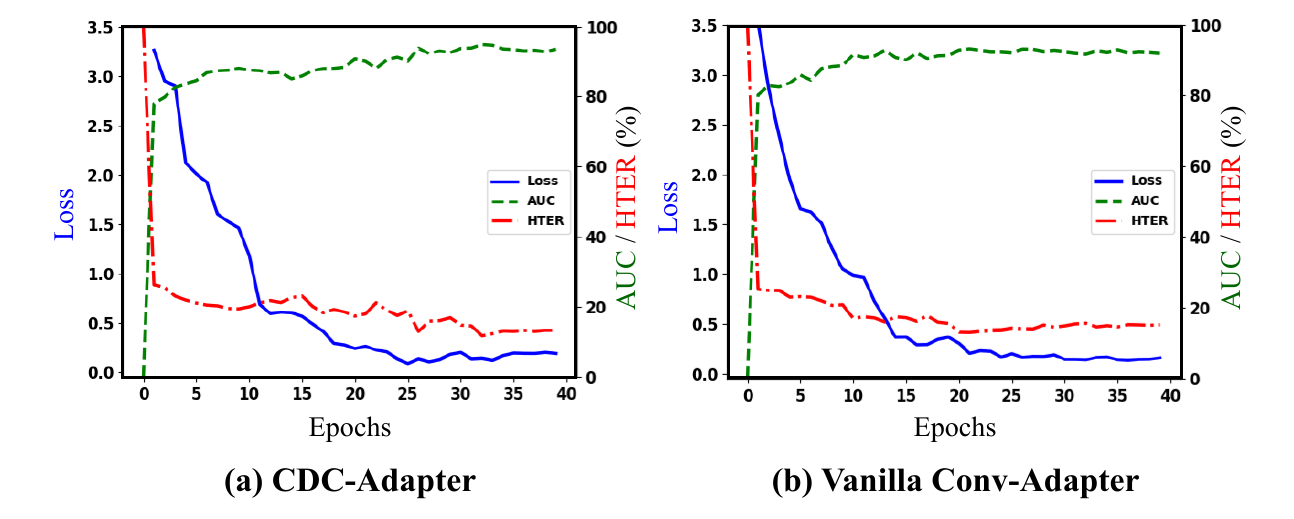}
    \vspace{-0.7cm}
    \centering
    \caption{Convergence speed of different convolutional Adapter. (a) CDC (Central Difference Convolution)-Adapter. (b) Vanilla Convolutional-Adapter.}
    \label{fig:CDC}
\vspace{-0.3cm}
\end{figure}

In Fig.~\ref{fig:CDC}, we compare the convergence speed of Adapters based on vanilla convolution and CDC. Combining the results from Tab. \ref{table: ablation study 3}, they demonstrate that the convergence speed of the vanilla convolution (which tends to stabilize after about 20 epochs) is faster then CDC (which tends to stabilize after about 30 epochs), and the performance of CDC is better. This phenomenon indicates that vanilla convolutional Adapter may have higher risk of overfitting compared to CDC-Adapter, CDC-Adapter is more robust for our backbone's fine-tuning.

\section{Algorithm}
\label{sec:algorithm}
The multi-modal PG-IRM algorithm is shown as below. Additionally, our input contains three modalities and includes a constraint term of angle margin in the total loss, therefore, the algorithm is designed with dual alignment of hyperplanes and angles. Our DADM optimization pipeline:

\begin{algorithm}[h!]
\renewcommand{\algorithmicrequire}{\textbf{Input:}}
\renewcommand{\algorithmicensure}{\textbf{Output:}}
\caption{The  optimization pipeline of \textbf{DADM}}
\label{alg:1}
\begin{algorithmic}[1]
\REQUIRE Source Data $S=\{x_i^{\mathrm{RGB}},x_i^{\mathrm{D}},x_i^{\mathrm{I}},y_j,e_i\}_i^N$,Target Data
$T=\{x_j^{\mathrm{RGB}},x_j^{\mathrm{D}},x_j^{\mathrm{I}},y_j\}_j^M$ , neural network $\phi(\cdot)$, classifiers $\beta_{e_1}, \beta_{e_2}, \cdots, \beta_{\mathcal{E}}$, learning rate $\gamma$, alignment parameter $\alpha$, alignment starting epoch $T_{\alpha}$.
\ENSURE $\phi(\cdot)$, $\mathrm{mean}(\beta_{e_1}, \beta_{e_2}, \cdots, \beta_{\mathcal{E}})$
\FOR {t in 0, 1, $\cdots$, T}
\STATE \textbf{Data Prep}: Sampling a mini-batch $B$ samples, $X_s=\{x_i^{\mathrm{RGB}},x_i^{\mathrm{D}},x_i^{\mathrm{I}},y_j,e_i\}^B_i$
\STATE \textbf{Forward}: Obtain multi-modal features and scores, $[f_i^{\mathrm{RGB}},f_i^{\mathrm{D}},f_i^{\mathrm{I}}]_{e_i}=\phi^t([x_i^{\mathrm{RGB}},x_i^{\mathrm{D}},x_i^{\mathrm{I}}]_{e_i}), \hat{y}_{e_i}= \beta_{e_i}^t [f_i^{\mathrm{RGB}},f_i^{\mathrm{D}},f_i^{\mathrm{I}}]_{e_i}$ 
\STATE \textbf{Backward}: Compute $\mathcal{L}_{\mathrm{total}}$, update $\phi^{t+1}=\phi^{t}-\gamma \nabla_{\phi^{t}}\mathcal{L}_{\mathrm{total}}$
\FOR {$e \in \mathcal{E}$}
\STATE $\tilde{\beta}_e^{t+1}=\beta_e^{t}- \gamma \nabla_{\beta_e^{t}}\mathcal{L}_{\mathrm{total}}$
\STATE select $\beta_{\bar{e}}^{t}$ with $\bar{e}=\underset{e'\in\mathcal{E} \backslash e}{\mathrm{argmax}}||\tilde{\beta}_e^{t+1}-\beta_{e'}^{t}||_2$
\STATE $\alpha'=1-\textbf{1}_{1>T_{\alpha}}(1-\alpha)$
\STATE $\beta_e^{t+1}=\alpha'\tilde{\beta}_e^{t+1}+(1-\alpha')\beta_{\bar{e}}^{t}$
\ENDFOR
\STATE $\bar{\beta}^{t+1}=\mathrm{mean}(\beta_{e_1}^{t+1}, \beta_{e_2}^{t+1}, \cdots, \beta_{\mathcal{E}}^{t+1})$
\STATE \textbf{Evaluate}: Test $\phi^{t+1}(\cdot)$, $\bar{\beta}^{t+1}$ on $T$ 
\IF {performance better} 
\STATE update $\phi^*(\cdot)=\phi^{t+1}(\cdot),\beta^*=\bar{\beta}^{t+1}$
\ENDIF
\ENDFOR
\end{algorithmic}  
\textbf{Return}  $\phi^*(\cdot)$, $\beta^*$
\end{algorithm}

\section{Proofs of the Necessity of Domain Alignment and Angle Alignment}
\label{sec:proofs of the necessity of domain alignment and angle alignment}
Invariant Risk Minimization (IRM) is a challenging bi-level optimization problem that is hard to solve. Thanks to the efforts of Sun et. al \cite{sun2023rethinking}, they propose the Projected Gradient Optimization for IRM (PG-IRM) which is an equivalent objective to IRM, with strict proof, and it is easier to optimize. The brief proof process is as follows:

\noindent
\textbf{Theorem 1. Projected Gradient Optimization IRM objective is equivalent to IRM objective.} \textit{For all $\alpha \in (0,1)$, the IRM objective is equivalent to the following objective:}
\begin{align}
\label{eqn:PG-IRM}
& \underset{\phi,\beta_{e_1}, \cdots, \beta_{\mathcal{E}}}{\mathrm{min}} \frac{1}{|\mathcal{E}|} \sum_{e \in \mathcal{E}} R^e(\phi,\beta_{e}), \notag \\
s.t. \, &   \forall e \in \mathcal{E}, \, \exists \beta_{e}\in \Omega_e(\phi), \beta_{e} \in \Upsilon
_{\alpha}(\beta_{e}) \, ,
\end{align}
\textit{where the parametric constrained set for each environment is simplified as $\Omega_e(\phi)=\underset{\beta}{\mathrm{argmin}}R^e(\phi,\beta)$, and the $\alpha$-adjacency set is defined as:}
\begin{align}
\label{eqn:a-adjacency set}
\Upsilon_{\alpha}(\beta_{e}) = &  \{v \, |\, \underset{e'\in\mathcal{E} \backslash e }{\mathrm{max}}\underset{\beta_{e'}\in\Omega_{e'}(\phi) }{\mathrm{min}}||v-\beta_{e'}||_2 \notag \\
& \leq \alpha \, \underset{e'\in\mathcal{E} \backslash e }{\mathrm{max}}\underset{\beta_{e'}\in\Omega_{e'}(\phi) }{\mathrm{min}}||\beta_e-\beta_{e'}||_2\}.
\end{align}

\noindent
\textbf{Proofs 1.} 

\noindent
The IRM objective is the following constrained optimization problem:
\begin{align}
\label{eqn:IRM supp}
& \underset{\phi,\beta^*}{\mathrm{min}} \frac{1}{|\mathcal{E}|} \sum_{e \in \mathcal{E}} R^e(\phi,\beta^*), \notag \\
s.t. \, & \beta^* \in \underset{\beta}{\mathrm{argmin}}R^e(\phi,\beta), \, \forall e \in \mathcal{E},
\end{align}
where $\phi$ represents a neural network, $\beta$ denotes the hyperplane for classification, $\mathcal{E}=\{e_1,e_2,\cdots,e_{|\mathcal{E}|}\}$ represents the entire environment, $e$ is one of the sub-environments, and $f(x;\beta,\phi)$ is the function processing $x$ via $\phi,\beta$ and obtaining $y$. The risk function $R^e(\phi,\beta)$, based on the loss function $\mathcal{L}(\cdot,\cdot)$, for a given environment $e$, is defined as:
\begin{align}
\label{eqn:risk function}
R^e(\phi,\beta) = \mathbb{E}_{(x,y)\sim
e}[\mathcal{L}(f(x;\beta,\phi),y)].
\end{align}
The constrain $\beta^* = \underset{\beta}{\mathrm{argmin}}R^e(\phi,\beta), \, \forall e \in \mathcal{E},$ means that the $\beta^*$ is the optimal linear classifier for all $e \in \mathcal{E}$, which is equivalent to $\beta^* \in \underset{e \in \mathcal{E}}{\cap} \Omega_e(\phi),$ and equivalent to:
\begin{align}
\label{eqn:constrain 0}
\forall e \in \mathcal{E}, \, \exists \beta_{e}\in \Omega_e(\phi), \beta^* = \beta_{e} \, .
\end{align}
This indicates that for all $e \in \mathcal{E}$, there is a hyperplane in the optimal set $\Omega_e(\phi)$ that also lies in the intersection of other environments' optimal set ($\underset{e'\in\mathcal{E} \backslash e}{\cap} \Omega_{e'}(\phi)$), i.e.:
\begin{align}
\label{eqn:constrain 1}
\forall e \in \mathcal{E}, \, \exists \beta_{e}\in \Omega_e(\phi), \beta_{e} \in \underset{e'\in\mathcal{E} \backslash e}{\cap} \Omega_{e'}(\phi) \, .
\end{align}
Sun et. al \cite{sun2023rethinking} relax the constrain to:
\begin{align}
\label{eqn:constrain 2}
\beta_{e} \in \underset{e'\in\mathcal{E} \backslash e}{\cap} \Omega_{e'}(\phi) \rightarrow \underset{e'\in\mathcal{E} \backslash e}{\mathrm{max}} ||\beta_{e} - \Omega_{e'}(\phi)||_2 \leq \epsilon \, ,
\end{align}
due to one key challenge for constrain \ref{eqn:constrain 1} is that there is a no guarantee that is non-empty for a feature extractor $\phi$ and $\beta_e$. Then they define the $l_2$ distance between a vector $\beta$ and a set $\Omega$ as: $||\beta-\Omega||_2=\underset{e'\in\mathcal{E} \backslash e}{\mathrm{min}}||\beta-\upsilon||_2$.
Practically, $\epsilon$ can be set to be any variable converging to 0 during the optimization stage. Without losing the generality, they change the constraint to the following form:
\begin{align}
\label{eqn:a-adjacency set 1}
& \forall e \in \mathcal{E}, \, \exists \beta_{e}\in \Omega_e(\phi), \\
 & \underset{e'\in\mathcal{E} \backslash e } {\mathrm{max}}\underset{\beta_{e'}\in\Omega_{e'}(\phi) }{\mathrm{min}}||\beta_e-\beta_{e'}||_2  \leq  \notag \\
&  \alpha \, \underset{e'\in\mathcal{E} \backslash e }{\mathrm{max}}\underset{\beta_{e'}\in\Omega_{e'}(\phi) }{\mathrm{min}}||\beta_e-\beta_{e'}||_2,
\end{align}
where $\alpha \in (0,1)$. Note that constraint \ref{eqn:a-adjacency set 1} will be satisfied only when $\underset{e'\in\mathcal{E} \backslash e }{\mathrm{max}}\underset{\beta_{e'}\in\Omega_{e'}(\phi) }{\mathrm{min}}||\beta_e-\beta_{e'}||_2=0$. Therefore constraint \ref{eqn:constrain 1} is equivalent to constraint \ref{eqn:a-adjacency set 1}.

Let's define $\Upsilon_{\alpha}(\beta_{e})$:
\begin{align}
\label{eqn:a-adjacency set 2}
\Upsilon_{\alpha}(\beta_{e}) = &  \{v \, |\, \underset{e'\in\mathcal{E} \backslash e }{\mathrm{max}}\underset{\beta_{e'}\in\Omega_{e'}(\phi) }{\mathrm{min}}||v-\beta_{e'}||_2 \notag \\
& \leq \alpha \, \underset{e'\in\mathcal{E} \backslash e }{\mathrm{max}}\underset{\beta_{e'}\in\Omega_{e'}(\phi) }{\mathrm{min}}||\beta_e-\beta_{e'}||_2\},
\end{align}
then the constraint \ref{eqn:constrain 0} can be simplified to: 
\begin{align}
\label{eqn:constraint 3}
s.t. \, &   \forall e \in \mathcal{E}, \, \exists \beta_{e}\in \Omega_e(\phi), \beta_{e} \in \Upsilon
_{\alpha}(\beta_{e}) \, .
\end{align}
\noindent
\textbf{Proofs 1 Completed}. 

Above Theorem 1 ensures that the PG-IRM's optimization objective being equivalent to the IRM's optimization objective.

\noindent
\textbf{Why we need dual alignment of hyperplane and angle margin?} In uni-modality scenarios, misalignment has always been a critical concern, as it relates to whether domain-invariant representations have been truly learned. MMDG \cite{lin2024suppress} found that directly incorporating multi-modality into DG-FAS can result in performance degradation, indicating the significance impact from domain and modality misalignment. Therefore, dual alignment of modality and domain is crucial.

\noindent
\textbf{Theorem 2. Misalignment of angle margin for modality features leads to severe shift and difficult convergence of the optimal classification hyperplane $\beta^*$ in PG-IRM.} \textit{For misaligned angle margin among modalities features in varies domains $[f^e_0, ... ,f^e_i, ...,f^e_\mathcal{M}]_\mathcal{M} \in \mathcal{E}$, where \([f^e_0, ... ,f^e_i, ...,f^e_\mathcal{M}]=\phi([x^e_0, ... ,x^e_i, ...,x^e_\mathcal{M}]) \in \mathbb{R}^{\mathcal{D} \times \mathcal{M}}\), $x^e_i$ represents single modality input $i$ from environment $e$. The the optimal classification hyperplane $\beta^*$ will severely shift.} 

\noindent
\textbf{Notation declarations}.

\noindent
For $f^e_i(k)$, $f$ represents the feature, the superscript $e$ denotes $f^e$ comes from environment $e$, the subscript $i$ denotes the $i$-th modality feature, $f(k)$ indicates the $k$-th element of $f$. Specially, $f^e$ (without subscript) denotes final fusion feature from environment $e$, $f^e(k)$ indicates the $k$-th element of $f^e$.

\noindent
\textbf{Proofs 2.} 

\noindent
For \([f^e_0, ... ,f^e_i, ...,f^e_\mathcal{M}]=\phi([x^e_0, ... ,x^e_i, ...,x^e_\mathcal{M}]) \in \mathbb{R}^{\mathcal{D} \times \mathcal{M}}\), where $f^e_i \in \mathbb{R}^{\mathcal{D} \times 1}$ and $\mathcal{M}$ is the number of modalities, we construct a modality matrix for environment $e$:
\begin{align}
\label{eqn:F}
\mathrm{F}^e = [f^e_0, ... ,f^e_i, ...,f^e_\mathcal{M}] \in \mathbb{R}^{\mathcal{D} \times \mathcal{M}},
\end{align}
the final fusion feature $f^e$ is obtained via a linear projecting \(\mathrm{P} \in \mathbb{R}^{\mathcal{M} \times 1}\):
\begin{align}
\label{eqn:f}
f^e =\mathrm{F}^e\mathrm{P} = \sum_i^\mathcal{M} p_i f^e_i \in \mathbb{R}^{\mathcal{D} \times 1}.
\end{align}

\noindent
\textbf{Intra-domain case.} 

\noindent
The intra-domain co-variance matrix of $f^e$ is as follows (for the simplicity, we omit the superscript $e$):
\begin{align}
\label{eqn:Var}
 \mathbb{E}[f] & = \sum_i^\mathcal{M} p(i) \mathbb{E}[f_i], \notag
\\
  \mathbb{D}[f] &  = \mathbb{E}[(f - \mathbb{E}[f])(f - \mathbb{E}[f])^{\textbf{T}}] \notag \\
& = \mathbb{E}[ff^{\textbf{T}}] - \mathbb{E}[f]\mathbb{E}[f]^{\textbf{T}} \notag \\
&  =\mathbb{E}[\mathrm{F}\mathrm{P}\mathrm{P}^{\textbf{T}}\mathrm{F}^{\textbf{T}}] - \mathbb{E}[f]\mathbb{E}[f]^{\textbf{T}},
\end{align}
where $p(i)$ denotes the $i$-th element of $\mathrm{P}$.

Assuming that the modality features $f^e_i$ have been normalized before classification by the classifier $\beta$, i.e., $\mathbb{E}[f_i]=0, ||f_i||=1,$ thus Eqn \ref{eqn:Var} can be rewritten as:
\begin{align}
\label{eqn:Var 1}
\mathbb{E}[f] & = 0, \notag \\
\mathbb{D}[f] & =  \mathbb{E}[\mathrm{F}\mathrm{P}\mathrm{P}^{\textbf{T}}\mathrm{F}^{\textbf{T}}],
\end{align}
and $k$-th diagonal elements of co-variance matrix $\mathbb{D}[f]$, which represents the $\mathbb{D}[f(k)]$:
\begin{align}
\label{eqn:D[f] element}
\mathbb{D}[f(k)] &  = p(k)^2\mathbb{E}[<f_k, f_k>] \notag \\
& = p(k)^2\mathbb{E}[||f_k||\cdot||f_k||\mathrm{cos}\theta_{k}] \notag \\
 & = p(k)^2\mathbb{E}[\mathrm{cos}\theta_{kk}],
\end{align}
where $<,>$ indicates the inner product, $p(k)$ denotes the $k$-th element of $\mathrm{P}$.

Since we consider that the distribution of angles $\theta_{kk}$ ($\theta$) without intervention generally does not approach a constant distribution, in order to maintain generality, we suppose that $\theta$ follows a Gaussian distribution with $\mu$ and variance $\sigma$, $N(\mu,\sigma)$:
\begin{align}
\label{eqn:Gaussian distribution}
f(\theta) = \frac{1}{\sigma \sqrt{2 \pi}} \mathrm{exp}(-\frac{(\theta - \mu)^2}{2\sigma^2} ).
\end{align}
Then we can calculate the value of $\mathbb{E}[\mathrm{cos}\theta]$:
\begin{align}
\label{eqn:expectation}
\mathbb{E}[\mathrm{cos}(\theta)] = \int_{- \infty}^{\infty} \mathrm{cos}(\theta) \cdot \frac{1}{\sigma \sqrt{2 \pi}} \mathrm{exp}(-\frac{(\theta - \mu)^2}{2\sigma^2} ) \mathrm{d}\theta,
\end{align}
\begin{align}
\label{eqn:expectation 1}
\mathrm{cos}(\theta) = \frac{ \mathrm{exp}(-i\theta)+\mathrm{exp}(i\theta)}{2},
\end{align}
\begin{align}
\label{eqn:expectation 2}
\mathbb{E}[\mathrm{cos}(\theta)] = \frac{1}{2}(\mathbb{E}[\mathrm{exp}(-i\theta)]+\mathbb{E}[\mathrm{exp}(i\theta)]).
\end{align}
For Gaussian distribution $N(\mu,\sigma)$, its characteristic function is $\Phi(t)=\mathbb{E}[\mathrm{exp}(-it\theta)]=\mathrm{exp}(i\mu t-\frac{\sigma^2t^2}{2})$.

The characteristic function when $t$ takes 1 and -1 is:
\begin{align}
\label{eqn:expectation 3}
\mathbb{E}[\mathrm{exp}(i\theta)] = \mathrm{exp}(i\mu-\frac{\sigma^2}{2}), \\
\label{eqn:expectation 4}
\mathbb{E}[\mathrm{exp}(-i\theta)] =  \mathrm{exp}(-i\mu-\frac{\sigma^2}{2}).
\end{align}
Substitute Eqn \ref{eqn:expectation 3}, \ref{eqn:expectation 4} into Eqn \ref{eqn:expectation 2}:
\begin{align}
\label{eqn:expectation 5}
\mathbb{E}[\mathrm{cos}(\theta)] & = \frac{1}{2}(\mathrm{exp}(i\mu-\frac{\sigma^2}{2})+\mathrm{exp}(-i\mu-\frac{\sigma^2}{2})), \notag \\
\mathbb{E}[\mathrm{cos}(\theta)] & = \frac{1}{2}\mathrm{exp}(-\frac{\sigma^2}{2})\cdot 2 \mathrm{cos}(\mu) = \mathrm{exp}(-\frac{\sigma^2}{2}) \mathrm{cos}(\mu).
\end{align}
Thus, increasing the variance ($\sigma$) of $\theta$ will leads to a decrease in the value of $\mathbb{D}[f(k)]$:
\begin{align}
\label{eqn:expectation 6}
\mathbb{D}[f(k)] = p(k)^2 \mathrm{exp}(-\frac{\sigma^2}{2})\mathrm{cos}(\mu).
\end{align}
This result indicates that when the angle margins $\theta$ between modalities exhibit a significant disturbance, the $\mathbb{D}[f(k)]$ will decrease.

\begin{figure*}[t!]
\centering
    \includegraphics[width=1.\textwidth]{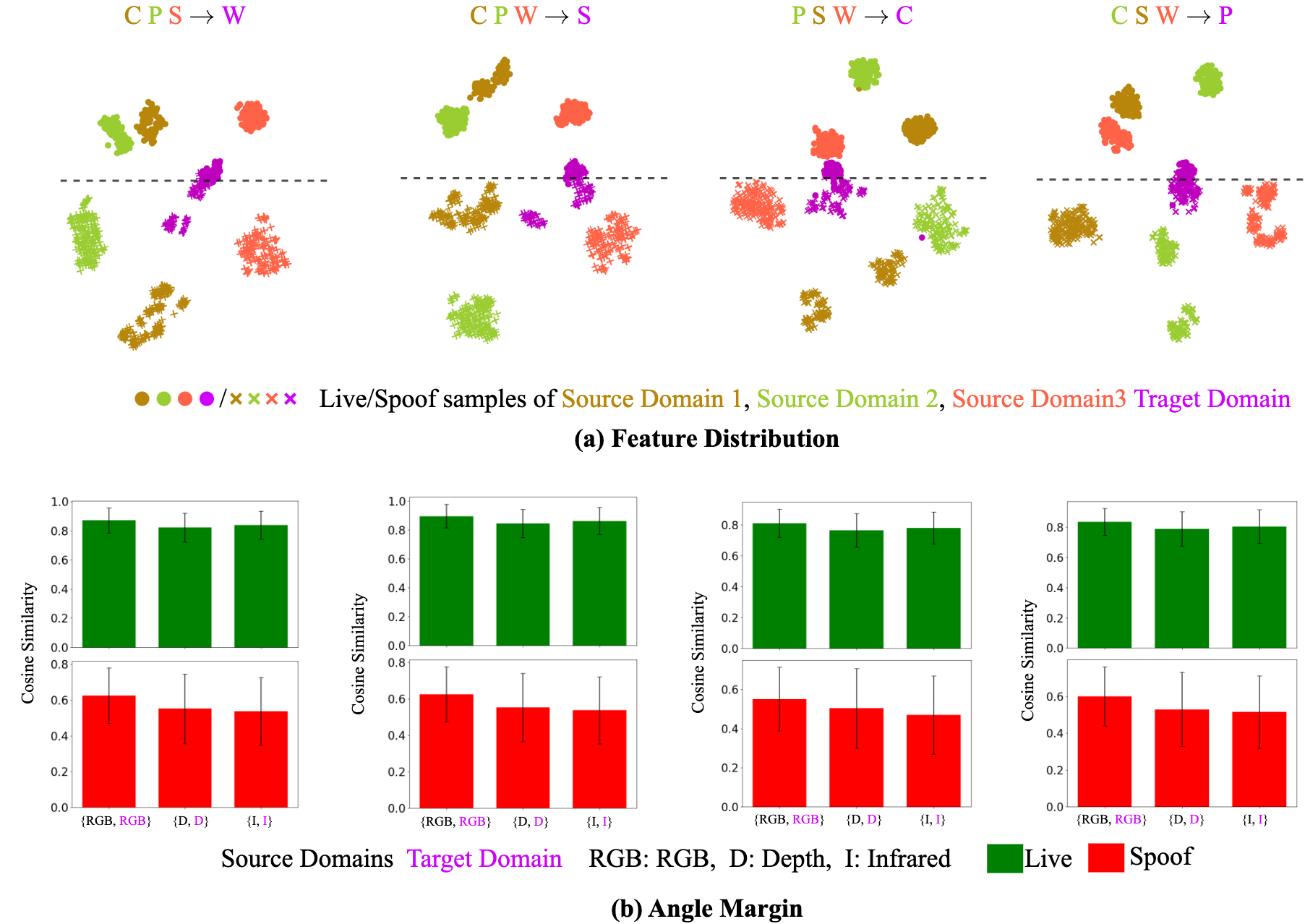}
    \caption{Illustration of dual alignment of domain and modality for four sub-protocols. (a) feature distribution of source and target domains, the dotted line represents the decision hyperplane in 2D space. (b) Mean and Std. of cosine similarity among modalities in the source and target domains.}
    \label{fig:alignment of hyperplane and angle}
\vspace{-0.3cm}
\end{figure*}

\noindent
\textbf{Inter-domain case.} 

\noindent
The inter-domain co-variance matrix between $f^{e_1}$ and $f^{e_2}$ is as follows:
\begin{align}
\label{eqn:Co-Var}
 \mathbb{E}[f^{e_1}] & = \sum_i^\mathcal{M} p(i) \mathbb{E}[f^{e_1}_i], \, \mathbb{E}[f^{e_2}] = \sum_i^\mathcal{M} p(i) \mathbb{E}[f^{e_2}_i], \notag \\
  \mathbb{C}[f^{e_1},f^{e_2}] &  = \mathbb{E}[(f^{e_1} - \mathbb{E}[f^{e_1}])(f^{e_2} - \mathbb{E}[f^{e_2}])^{\textbf{T}}] \notag \\
& = \mathbb{E}[f^{e_1}f^{e_2\textbf{T}}] - \mathbb{E}[f^{e_1}]\mathbb{E}[f^{e_2}]^{\textbf{T}} \notag \\
&  =\mathbb{E}[\mathrm{F}^{e_1}\mathrm{P}^{e_1}\mathrm{P}^{e_2\textbf{T}}\mathrm{F}^{e_2\textbf{T}}] - \mathbb{E}[f^{e_1}]\mathbb{E}[f^{e_2}]^{\textbf{T}}.
\end{align}
Please note that $f^{e_1}$ and $f^{e_2}$ exhibit the same liveness label.

Assuming that the modality features $f^e_i$ have been normalized before classification by the classifier $\beta$, i.e., $\mathbb{E}[f^e_i]=0, ||f^e_i||=1,$ thus Eqn \ref{eqn:Co-Var} can be rewritten as:
\begin{align}
\label{eqn:Co-Var 1}
\mathbb{E}[f^{e_1}] & = 0, \, \mathbb{E}[f^{e_2}] = 0, \notag \\
  \mathbb{C}[f^{e_1},f^{e_2}] & =\mathbb{E}[\mathrm{F}^{e_1}\mathrm{P}^{e_1}\mathrm{P}^{e_2\textbf{T}}\mathrm{F}^{e_2\textbf{T}}],
\end{align}
the $k$-th diagonal elements of $\mathbb{C}[f^{e_1},f^{e_2}]$, which represents the co-variance between $f^{e_1}(k)$ and $f^{e_2}(k)$:
\begin{align}
\label{eqn:Var 2}
\mathbb{C}[f^{e_1}(k),f^{e_2}(k)] & = p(k)^2\mathbb{E}[<f^{e_1}_k,f^{e_2}_k>] \notag \\
& = p(k)^2\mathbb{E}[||f^{e_1}_k||\cdot||f^{e_2}_k||\mathrm{cos}\theta_{kk}] \notag \\
& = p(k)^2\mathbb{E}[\mathrm{cos}\theta_{kk}].
\end{align}
Similarly, we can also conclude that increasing the variance ($\sigma$) of $\theta$ will also lead to a decrease in the value of $\mathbb{C}[f^{e_1}(k),f^{e_2}(k)]$ according to \textbf{Intra-domain case}.

\noindent
\textbf{The impact of $\mathbb{D}[f(k)]$ on the convergence and shift of $\beta$}.

\noindent
Before computing the loss function, we need to use a linear classifier $\beta$ and $\mathrm{softmax}(\cdot)$ projecting final fusion feature $f$ to logits $\hat{y}$, where $z=\beta f$, $\hat{y}=[\frac{\mathrm{exp}(z_p)}{\mathrm{exp}(z_p)+\mathrm{exp}(z_n)}, \frac{\mathrm{exp}(z_n)}{\mathrm{exp}(z_p)+\mathrm{exp}(z_n)}]$, $p$ represents positive sample, $n$ represents negative sample. Considering that using cross-entropy loss:
\begin{align}
\label{eqn:cross-entropy}
\mathcal{L} = - \mathbb{I}(\small{\mathrm{label}_\mathrm{GT}})\hat{y} \mathrm{log}\hat{y}
\end{align}
where $\hat{y}_p, \hat{y}_n \in (0, 1)$, the gradient of $\mathcal{L}$:
\begin{align}
\label{eqn:gradient}
\nabla_{\beta_e^{t}}\mathcal{L} & = \mathbb{I}(\small{\mathrm{label}_\mathrm{GT}})\nabla_{\beta_e^{t}}\hat{y} \, \frac{\partial \mathcal{L}}{\partial \hat{y}} \notag \\
& = - \mathbb{I}(\small{\mathrm{label}_\mathrm{GT}}) \nabla_{\beta_e^{t}}\hat{y}(\mathrm{log}\hat{y} + 1)
\end{align}
then we consider the variance of $z_p=\beta_p f=\sum_k^\mathcal{D} w(k) f(k)$ (the same applies to the analysis of $z_n=\beta_n f$), which $\mathbb{D}[z_p] =\sum_k^\mathcal{D} w(k)^2 \mathbb{D}[f(k)]$, $w(k)$ is the weight of $\beta_p$. When the $\sigma$ of $\theta_{kk}$ increases, the $\mathbb{D}[f(k)]$ decreases, so does the $\mathbb{D}[z_p]$. The smaller $\mathbb{D}[z_p]$ and $\mathbb{D}[z_n]$ will lead to the difference between $z_p$ and $z_n$ may be subtle at the start of training (supposing that randomly initialization does not favor either $z_p$ or $z_n$), resulting in a flatten value of softmax output $\hat{y}$, i.e., the value of logits ($\hat{y}$) tend towards a uniform distribution. And we can easily know that the function $\hat{y}\mathrm{log}\hat{y}$ has its maximum value when the probability of $\hat{y}$ reaches $1/n$ (for two categories, $n$ equals to 2).

At this point, the drastic fluctuation in $\theta_{kk}$ will cause the absolute value of gradient $|\nabla_{\beta_e^{t}}\mathcal{L}|$ to be difficult to converge to a smaller value. According to line 6-9 in Algorithm \ref{alg:1}:
\begin{align}
\label{eqn:gradient shift}
& \beta_e^{t+1}=\alpha'\tilde{\beta}_e^{t+1}+(1-\alpha')\beta_{\bar{e}}^{t} \rightarrow \bar{e}=\underset{e'\in\mathcal{E} \backslash e}{\mathrm{argmax}}||\tilde{\beta}_e^{t+1}-\beta_{e'}^{t}||_2, \notag \\
& \beta_e^{t+1}=\alpha'(\beta_e^{t}- \gamma \nabla_{\beta_e^{t}}\mathcal{L}_{\mathrm{total}})+(1-\alpha')\beta_{\bar{e}}^{t}, \notag \\
& \beta_e^{t+1}= \beta_{\bar{e}}^{t} + \alpha'(\beta_e^{t} - \beta_{\bar{e}}^{t}) - \alpha' \gamma \nabla_{\beta_e^{t}}\mathcal{L}_{\mathrm{total}},
\end{align}
$t$ starts from 0 to T, the single-step shift will accumulate increasingly, resulting in the optimal classification hyperplane $\bar{\beta}^{T+1}=\mathrm{mean}(\beta_{e_1}^{T+1}, \beta_{e_2}^{T+1}, \cdots, \beta_{\mathcal{E}}^{T+1})$ shift severely.

\noindent
\textbf{Proofs 2  Completed.} 

\section{Visualization}
\label{sec:visualization}
\noindent
\textbf{Comprehensive visualization of dual alignment of domain and modality}. Fig. \ref{fig:alignment of hyperplane and angle} presents the visualizations of dual alignment of hyperplanes and angles for four sub-protocols in Tab. \ref{table: protocol 1}. we can observe that in the sub-protocols CPS $\rightarrow$ W and CPW $\rightarrow$ S, the hyperplane for the live/spoof decision remains consistent across different source domains and is also transferable to unseen target domain. Moreover, the angles between the source domains and the target domain are relatively close to the expected values. In contrast, the other two sub-protocols PSW $\rightarrow$ C and CSW $\rightarrow$ P, exhibit slightly poorer illustration. Correspondingly, they also show poorer performance in Tab. \ref{table: protocol 1}, which might be due to their encountering of a more significant domain shift.

\begin{figure}[t!]
\centering
    \includegraphics[width=0.48\textwidth]{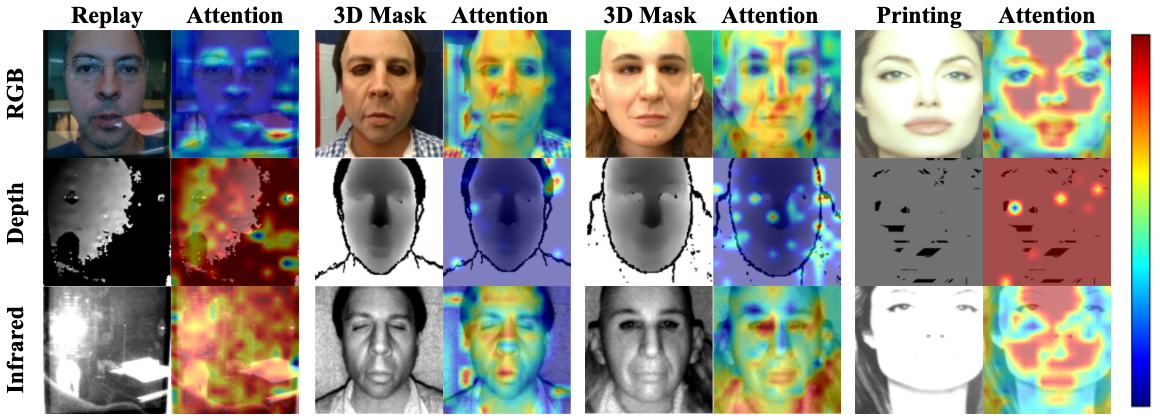}
   \vspace{-0.7cm}
    \caption{More visualization attention maps on varies attack samples, for example, replay attack, 3D mask, paper printing.}
    \label{fig:more visualization attention maps}
\vspace{-0.1cm}
\end{figure}

\noindent
\textbf{More visualization attention maps.} In Fig. \ref{fig:more visualization attention maps}, For 3D masks, the face region in the depth map is shown with cooler color, indicating its weak influence. For paper-printing attacks, depth information is particularly revealing of spoof cues, thereby warranting higher importance. For video replay attacks, more obvious spoofing traces were observed from infrared and depth maps, so both of them have higher importance than RGB.

\end{document}